\documentclass[journal]{IEEEtran}
\ifCLASSOPTIONcompsoc
\usepackage[nocompress]{cite}
\else
\usepackage{cite}
\fi

\usepackage{times}
\usepackage{epsfig}
\usepackage{graphicx}
\usepackage[ruled,linesnumbered]{algorithm2e}
\usepackage{amsmath}
\usepackage{amssymb}
\usepackage{subfigure}
\usepackage{setspace}
\usepackage{multirow}
\usepackage{cite}
\usepackage{comment}
\usepackage{booktabs}
\usepackage{color}
\usepackage{booktabs}
\usepackage[normalem]{ulem}
\usepackage[switch, displaymath, mathlines]{lineno}
\newcommand\zxhzxh[1]{\textcolor{black}{#1}}

\newcommand\XP[1]{\textcolor{black}{#1}}

\newcommand\zxh[1]{\textcolor{black}{#1}}

\begin{document}
	
	% \title{Global Precedence plus Local Attention : Modulating Dual-Pathway Vision Transformer Module for Image Classification}
	\title{Deep Class Incremental Learning  from Decentralized Data}
	%\title{Class Incremental Learning of Deep Networks from Decentralized Data}
	
	\author{Xiaohan~Zhang,
		Songlin~Dong,
		Jinjie~Chen,
		Qi~Tian~\IEEEmembership{Fellow,~IEEE},
		~Yihong~Gong~\IEEEmembership{Fellow,~IEEE,} and Xiaopeng~Hong~\IEEEmembership{Member,~IEEE}%
		\thanks{The manuscript was submitted on August 31st, 2021 and revised on March 11th, 2022. Xiaohan Zhang and Xiaopeng~Hong are with the School of Cyber Science and Engineering, Xi’an Jiaotong University;  Songlin Dong are with the College of Artificial Intelligence, Xi’an Jiaotong University; Jinjie Chen and Yihong~Gong are with the College of Software Engineering, Xi’an Jiaotong University, Xi'an 710049, China; Qi~Tian is with Cloud \& AI, Huawei Technologies, China.
			(e-mail: {zxh980111, dsl972731417, chenjinjie}@stu.xjtu.edu.cn;  ygong@mail.xjtu.edu.cn; hongxiaopeng@mail.xjtu.edu.cn; tian.qi1@huawei.com).
			}% <-this % stops a space
			%$^\dagger$Corresponding author: Xiaopeng Hong.}% <-this % stops a space
		\thanks{}}% <-this % stops a space
	
	\maketitle
	
	\markboth{SUBMITTED TO IEEE TRANSACTIONS ON NEURAL NETWORKS AND LEARNING SYSTEMS. REVISED VERSION.}%
	{Zhang \MakeLowercase{\textit{et al.}}: Class Incremental Learning of Deep Networks from Decentralized Data}
	
	
	
	\begin{abstract} 
		In this paper, we focus on a new and challenging decentralized machine learning paradigm in which there are continuous inflows of data to be addressed and the data are stored in multiple repositories. 
		We initiate the study of data decentralized class-incremental learning (DCIL) by making the following contributions. 
		Firstly, we formulate the DCIL problem and develop the experimental protocol. 
		Secondly, we introduce a paradigm to create a basic decentralized counterpart of typical (centralized) class-incremental learning approaches, and as a result, establish a benchmark for the DCIL study. 
		Thirdly, we further propose a Decentralized Composite knowledge Incremental Distillation framework (DCID) to transfer knowledge from historical models and multiple local sites to the general model continually. 
		% \zxh{We validate our DCID with three CIL methods against three existing decentralized learning methods.}
		DCID consists of three main components namely local class-incremental learning, collaborated knowledge distillation among \emph{local} models, and aggregated knowledge distillation from \emph{local} models to the \emph{general} one. We comprehensively investigate our DCID framework by using different implementation of the three components. Extensive experimental results demonstrate the effectiveness of our DCID framework. {The codes of the baseline methods and the proposed DCIL will be released at https://github.com/zxxxxh/DCIL.}
		%\zxh{We build the DCIL baselines by adapting three state-of-the-art decentralized learning methods to this new problem and compare our DCID with them under three CIL methods.}
		%Experiments on two public datasets demonstrate the effectiveness of our DCID framework.
	\end{abstract}
	
	\IEEEpeerreviewmaketitle
	\section{Introduction}\label{introduction}
	\IEEEPARstart{D}eep models has achieved great success in a wide range of artificial intelligence research fields~\cite{2016Deep,2017Fully,krizhevsky2012imagenet,ren2015faster,liu2017sphereface,chang2020devil,9511096}. Nevertheless, they have been shown to prone to the \emph{catastrophic forgetting} problem~\cite{mccloskey1989catastrophic}.
	\emph{Catastrophic forgetting} refers to the phenomenon where the performance of the deep model degrades seriously when evolving the model for new data. In response to this urgent problem, incremental learning (IL)~\cite{cauwenberghs2001incremental,kuzborskij2013n,lee2017overcoming,tao2020few,zhao2021video,liu2020more,DBLP:journals/corr/abs-2006-15524}, \emph{a.k.a.}, continual learning~\cite{2018Continual,tao2020bi,shin2017continual,zenke2017continual}, which is targeted at learning continuous incoming data streams while getting away with \emph{catastrophic forgetting}, has drawn increasing attention.
	
	\zxhzxh{Current incremental learning framework requires the 
		deep neural network models to process continuous streams of information in a centralized manner. 
		Despite its success, we argue that such a centralized setting is often impossible or impractical.
		More and more data emerge from and exist in ``isolated islands", which may be subject to various regularization or requirements in privacy. It is not always allowed to move data and use data out of their owners. 
		In addition, continuous inflows lead to a huge amount of data located in different repositories, which may cause huge communication and computational burden in bringing them together into a single repository for learning. }

	\zxhzxh{Therefore, it is crucial to enable learning models to be deployed in scenarios where data are located in different places and the learning process to be performed across time beyond the bounds of a single repository.
		Nevertheless, no existing machine learning paradigm such as the incremental learning and distributed learning are able to handle such complex scenarios and thus leave us a big challenge, as illustrated by Table~\ref{DL_CIL_DCIL}. In incremental learning {\cite{cauwenberghs2001incremental,kuzborskij2013n,lee2017overcoming,tao2020few,zhao2021video,2017iCaRL,tao_eccv20,Hou_2019_CVPR}}, a model is updated given a data stream continually coming from one \emph{single} repository. On the contrary, in Distributed Learning (DL) and Federated Learning~(FL) \cite{2018Federated,2019SCAFFOLD}, multiple models learnt by different repositories are aggregated to a \emph{general} model. Clearly, IL cannot process data from multiple repositories, while DL and FL cannot provide a mechanism to handle continuous data streams.}
	
	%In incremental learning \XPC{\cite{cauwenberghs2001incremental,kuzborskij2013n,lee2017overcoming,tao2020few,zhao2021video,2017iCaRL,tao_eccv20,Hou_2019_CVPR}}, a model is updated given a data stream of different classes continually coming from one \emph{single} repository. On the other hand, in Distributed Learning (DL) and Federated Learning~(FL) \XPC{\cite{2018Federated,2019SCAFFOLD}} multiple models learnt by different repositories are aggregated to a \emph{general} model. Clearly, DL and FL don't provide a mechanism to handle continuous data streams.}
	
	\begin{figure}[ht]
		\includegraphics[width=0.49\textwidth]{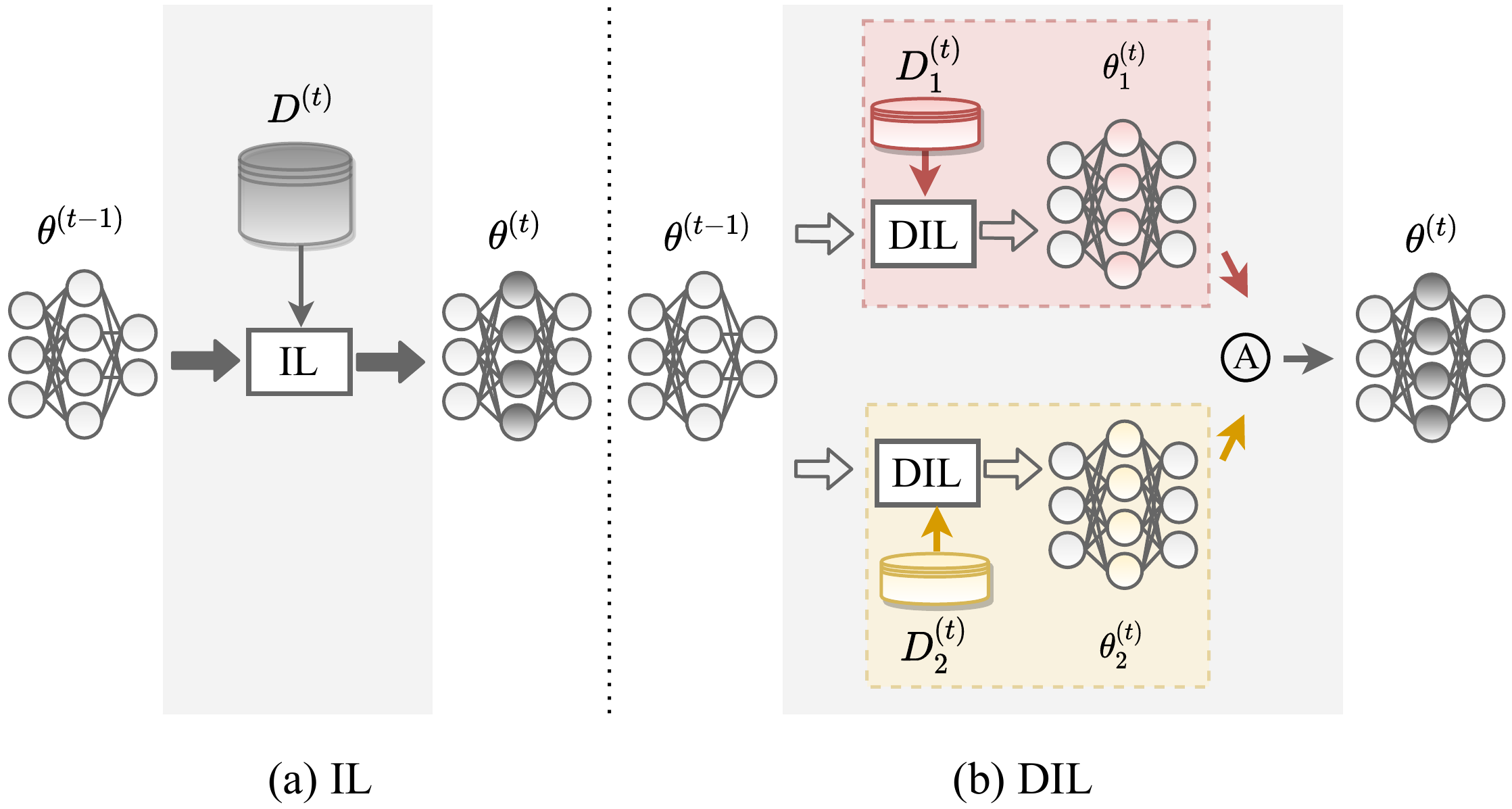}
		\caption{Comparisons of (a) the traditional deep (class-) Incremental Learning (IL) and (b) the Decentralized deep (class-) Incremental Learning paradigms (DIL). The area in gray indicates the current session of learning. \textcircled{A} refers to the function of model aggregation.}
		\label{fig:ildil}
	\end{figure}
	
	\begin{table}[tbp]
		\centering  
		\caption{{Different nature of data sources among traditional centralized class incremental learning~(CIL), {Distributed} learning~(DL) and Federated Learning (FL), and our proposed decentralized class incremental learning~(DCIL).}}
		\label{DL_CIL_DCIL}
		\renewcommand\arraystretch{1.2}
		\renewcommand\tabcolsep{2.5pt}
		\footnotesize
		\begin{tabular}{cccc}
			\hline\hline
			& CIL & DL / FL & DCIL \\ \hline
			data from multiple repositories & & $\checkmark$ & $\checkmark$ \\ 
			continuous data streams & $\checkmark$&	 & $\checkmark$\\ \hline\hline
		\end{tabular}
	\end{table}
	
	\XP{In this paper, we raise the concern about} such a new challenging scenario, where deep incremental learning shall be performed in a decentralized manner, as illustrated in Figure~\ref{fig:ildil}. 
	To meet this challenge, it is required to enable deep learning models to learn from new data residing on local sites like end devices, and in turn to promote the performance of the \emph{general} model on the main site continually. 
	%Take Smart Family Photo Album as an example, where a considerable number of photos can be token on occasion separately by different smart phones. A pre-trained CNN model is deployed in a central console as a \emph{general} model shared by the family members. To process and analyze the inflows of photos in real-time, a group of \emph{local} models is deployed and maintained in each smartphone. As the number of photos may increase rapidly, the local CNN models are required to learn from and adapt to newly emerging photos, which may distinctly vary from historical ones. Once local models are updated, the \emph{general} model shall communicate with the local ones, learn from them, and updates itself accordingly, without copying all users' photos in a single repository. The decentralized incremental learning algorithm can also be used in the field of multi-robot collaboration, such as welcome robots that can recognize human identities.\
	\zxhzxh{
		Take Smart Family Photo Album as an example, where a considerable number of photos can be token on occasion separately by different smart phones. A pre-trained CNN model is deployed in a central console as a \emph{general} model shared by the family members. To process and analyze the inflows of photos in real-time, a group of \emph{local} models is deployed and maintained in each smartphone. As the number of photos may increase rapidly, the local CNN models are required to learn from and adapt to newly emerging photos, which may distinctly vary from historical ones. Once local models are updated, the \emph{general} model shall communicate with the local ones, learn from them, and updates itself accordingly, without copying all users' photos in a single repository. The decentralized incremental learning algorithm can also be used in the field of multi-robot collaboration, such as welcoming robots that can recognize human identities. Each welcoming robot learns to identity guests over time and learn to recognize new guests without forgetting previous guests even if data of previous guests is unavailable. All welcoming robots can exchange models rather than data with guest privacy to each other in order to generate a main model that can recognize all guests.}
	% XP{Nevertheless, existing machine learning paradigm cannot handle these challenges. \zxh{In the Class incremental learning \XPC{references needed} describes a learning scenario where a model trains by data stream of different classes continually coming from one single repository. Distributed learning (DL) and Federated learning~(FL) \XPC{references needed} tackles the issue that aggregating multiple models training by different repositories to a \emph{general} model. }}
	
	%Federated learning~(FL) is an effective data decentralized learning approach that facilitates real-world applications.

	In response to this demand, we study the paradigm of \emph{decentralized class incremental learning} (DCIL). 
	%raise the concern about the decentralized deployment of class incremental learning paradigm and 
	%\zxh{The existing class incremental learning paradigm describes a learning scenario where a model trains by data stream of different classes continually coming from one single repository. Distributed learning paradigm tackles the issue that aggregating multiple models training by different repositories to a \emph{general} model. Federated learning~(FL) is an effective data decentralized learning approach that facilitates real-world applications. We compare the differences among centralized class incremental learning~(CIL), decentralized learning and our proposed \emph{decentralized class incremental learning} in Table~\ref{DL_CIL_DCIL}.}
	In DCIL, \XP{it is avoided to upload the data in local sites to the main site. It thus becomes \zxh{challenging to learn a general model with such limited information.}} 
	%local data in each local sites need not be uploaded to the main site, and only local models' parameters are sent to the main site.
	%\zxh{It is thus challenging to learn a general model with such limited information as good as the model obtained by centralized CIL. None of the existing CIL paradigms can solve this problem.}
	To kick off relevant research, we define a rationale DCIL learning and evaluation protocol on mainstream class-incremental-learning datasets. Moreover, under the protocol, we develop a DCIL paradigm to transform typical (centralized) class incremental learning approaches to their corresponding decentralized counterparts and built baseline DCIL results. 
	
	\begin{figure}[tb] 
		\centering  
		\subfigtopskip=2pt 
		\subfigcapskip=-2pt %设置子图与子标题之间的距离
		\subfigure[local model 0]{
			\label{reason.u0}
			\includegraphics[width=0.31\linewidth]{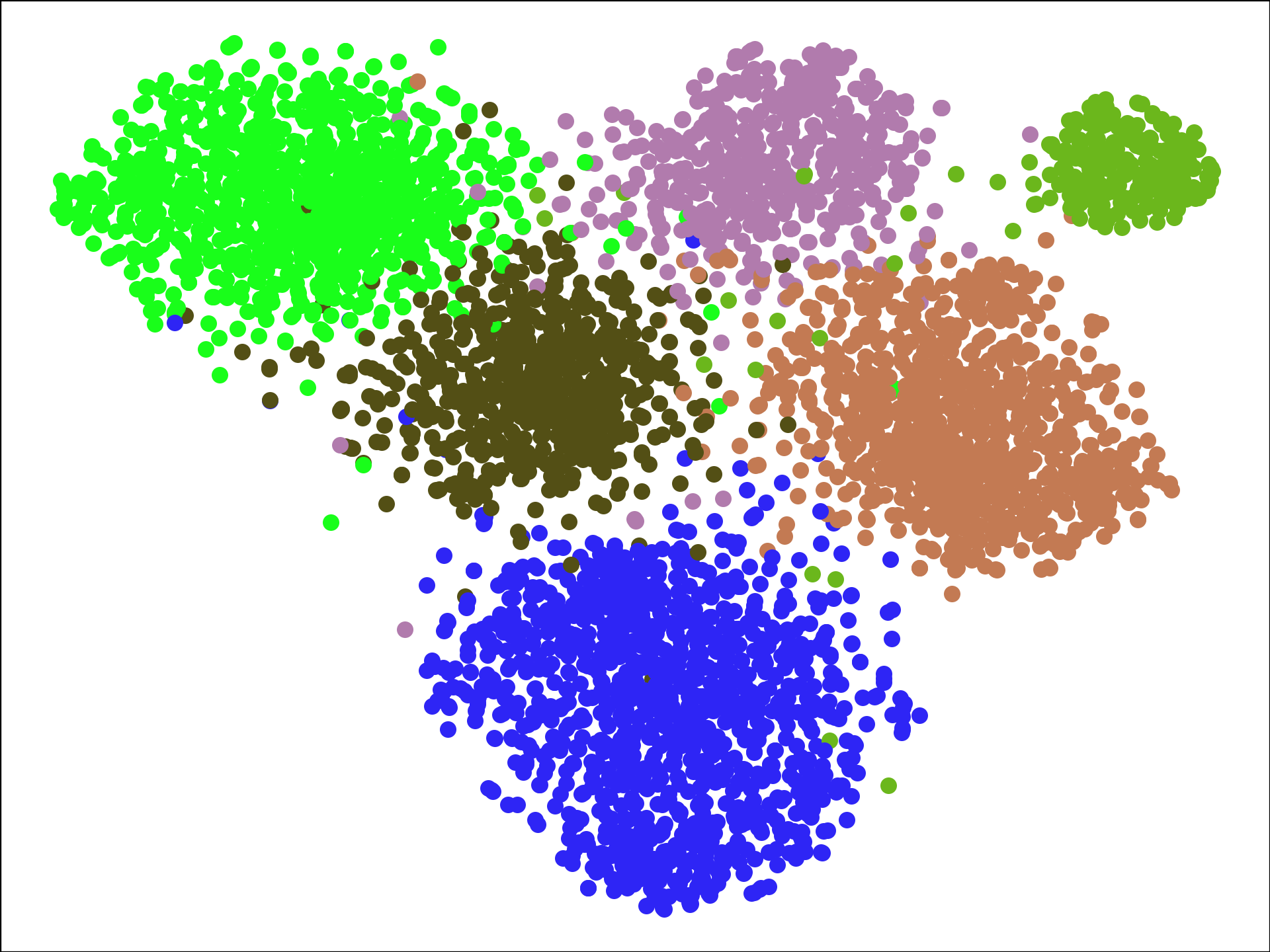}}
		\subfigure[local model 1]{
			\label{reason.u1}
			\includegraphics[width=0.31\linewidth]{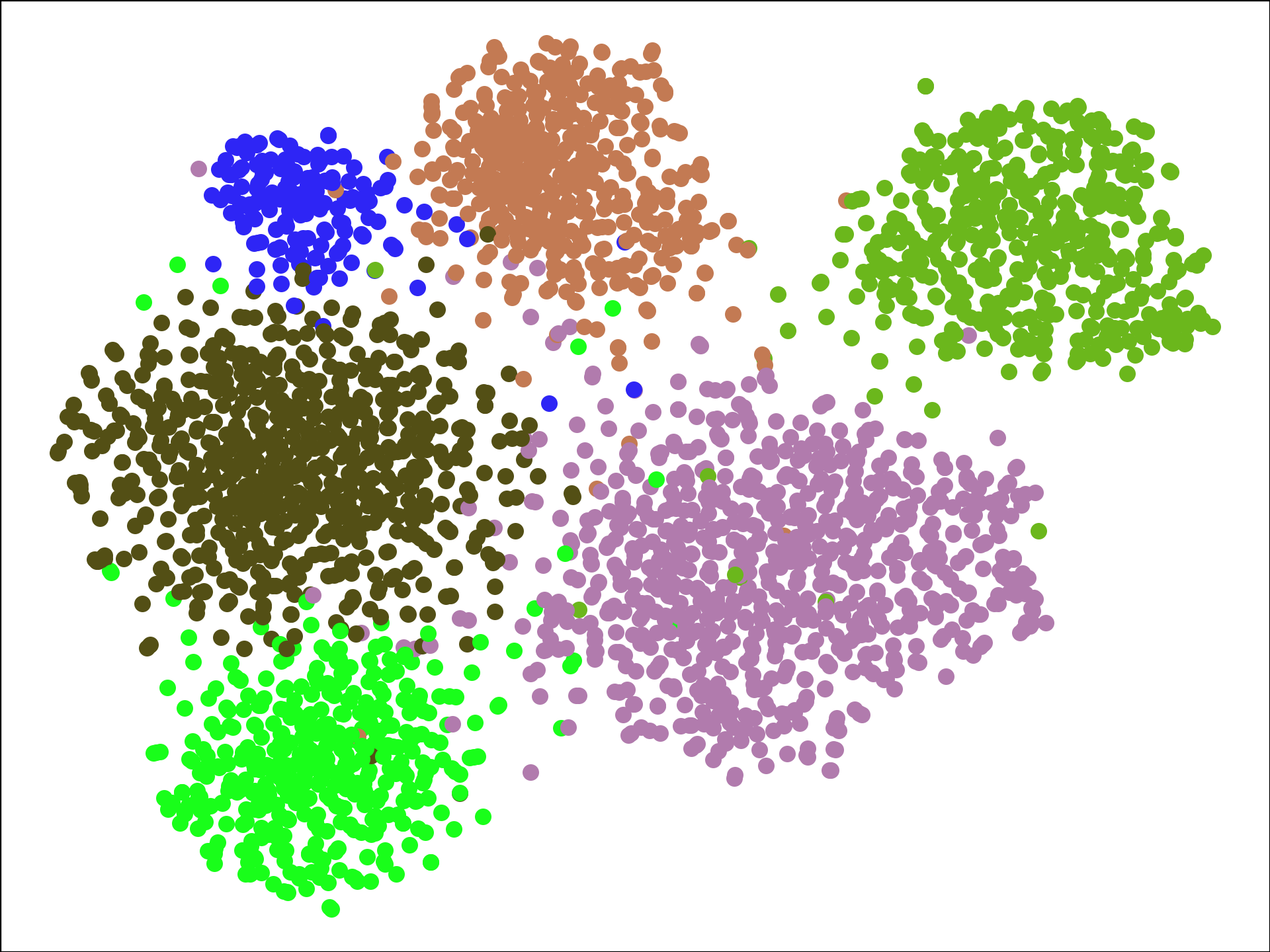}}
		\subfigure[general model]{
			\label{reason.avg}
			\includegraphics[width=0.31\linewidth]{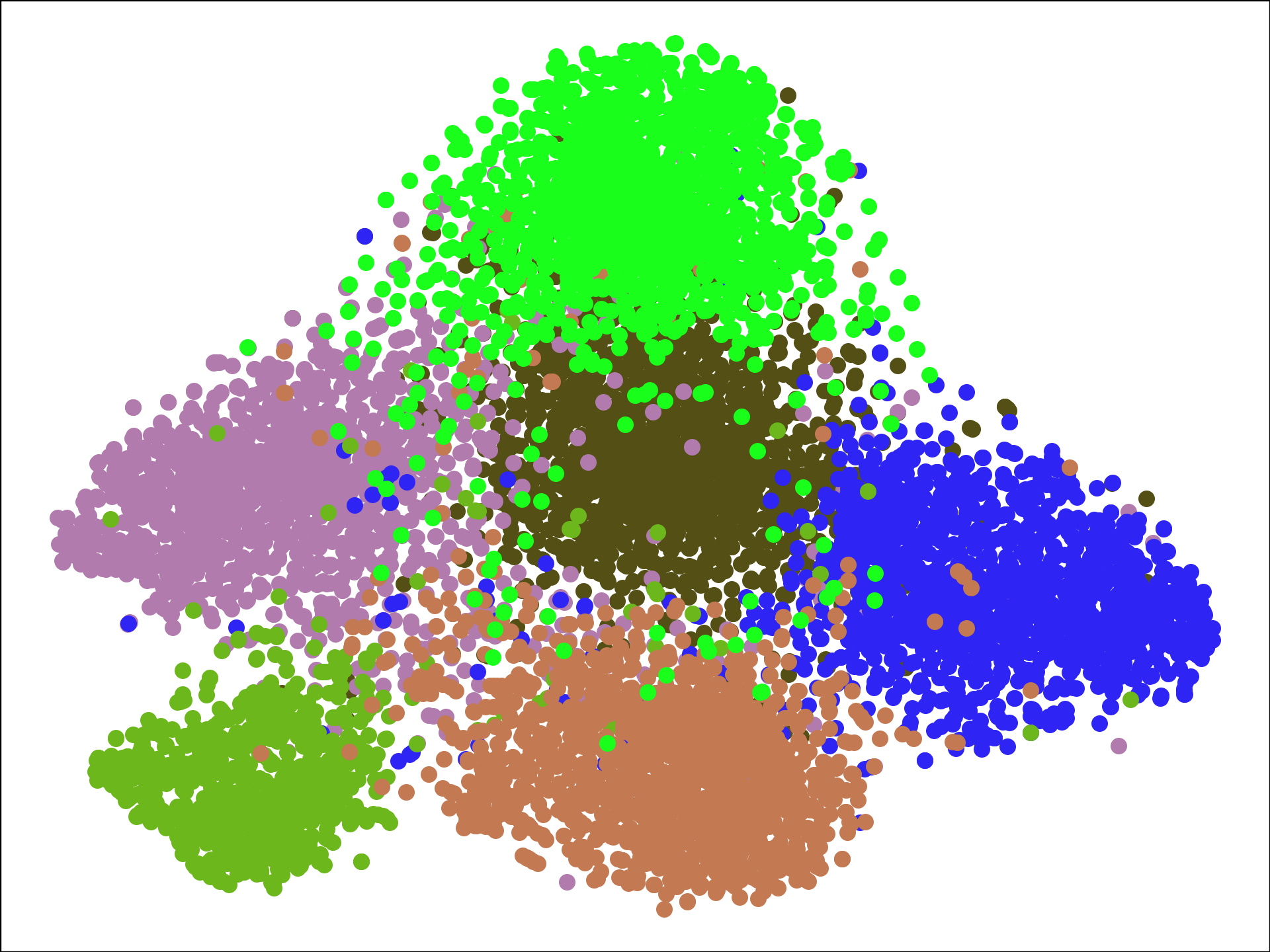}}
		\caption{\XP{Visualization of the phenomenon of \emph{model drift} after simple averaging with a basic DCIL setting using \emph{t}-SNE. An example of a DCIL model aggregated from two local sites on subImageNet is shown, where the number of base classes and the number of new classes in current session are 50 and 6 respectively. (a) and (b): the feature space of the two local models individually trained; (c): the feature space of the general model where noisy decision boundaries on the entire set of training samples at the current session is observed.}}
		\label{drifted_reason}
	\end{figure}
	
	Along with its promising prospect, there are still several problems of DCIL. On the one hand, as data are distributed among multiple repositories and streams, local sites can only access a portion of the entire data set, which inherently induces deflected local optimum during data decentralized learning~\cite{2019SCAFFOLD}. On the other hand, directly averaging the local model weights to form a general model may have detrimental effects on the performance of the general model. One reason may lie in the permutation-invariant property of neural network parameters, as one neural net has multiple equivalent counterparts with different orders of the same parameters~\cite{2019Bayesian}. As a result, 
	a phenomenon of \XP{\emph{model drift on the general model}} may occur, as shown in Figure~\ref{drifted_reason}.

	\XP{To further solve the challenge in DCIL and the limitations of the basic DCIL paradigm,} we propose a novel DCIL framework, termed \underline{D}ecentralized \underline{C}omposite  \underline{I}ncremental \underline{D}istillation framework (DCID) that enables to learn knowledge across time and multiple repositories. \XP{There are three main steps in the proposed DCID framework. Firstly, we introduce a data decentralized learning mechanism to perform class incremental learning. Secondly, we propose a collaborated knowledge distillation method to exchange knowledge among \emph{local} models so that \emph{local} models can self-consistently evolve, without the supervision from the general model. Finally, we design an aggregated knowledge distillation method to transfer knowledge from multiple \emph{local} models to update the \emph{general} model. The proposed approach outperforms the baseline DCIL algorithms, with a communication cost at the same level.}
	
	Briefly, the contributions of this paper are manifold:
	\begin{itemize}
		% \item We raise the concern on class incremental learning deployed in data decentralized, multi-agent scenarios.
		\item \XP{We recognize the importance and initiate the study of \emph{decentralized class incremental learning}~(DCIL).} Compared with the popularly studied \emph{class incremental learning}~(CIL), the problem setting of DCIL is more practical and challenging.
		%of class incremental learning deployed in data decentralized, multi-agent scenarios, and define the problem setting named \emph{decentralized class incremental learning}~(DCIL). Compared with the popularly studies \emph{class incremental learning}~(CIL), DCIL is more practical and  challenging.}
		%\item We establish a benchmark and \zxh{adapt the state-of-the-art decentralized learning methods} to provide the baseline results for the decentralized class incremental learning~(DCIL) study.
		\item We {propose a basic DCIL paradigm to decentralize state-of-the-art class incremental learning approaches and provide baseline results for the DCIL study.}
		%establish a benchmark and \zxh{adapt the state-of-the-art decentralized learning methods} to provide the baseline results for the decentralized class incremental learning~(DCIL) study.
		\item We propose a decentralized deep class incremental learning framework DCID, which consistently outperforms the baselines \XP{under various settings.}
		%under both IID and non-IID settings.
		%, regardless of particular CIL methods used. %\zxh{for both IID and non-IID settings.}
		% \item \zxh{The experiment results demonstrate the effectiveness of the proposed \emph{DCID} with three representative CIL methods 
	\end{itemize}
	
	\section{Related Work}\label{relatework}
	%\vspace{1mm}
	This study is relevant to class incremental learning and distributed learning / federated learning.
	\subsection{\zxhzxh{Deep neural networks} }
	
	\zxhzxh{Deep neural networks~(DNNs) have shown great ability to represent highly complex  functions. Deep neural networks, especially deep convolutional neural networks~(CNNs)~\cite{krizhevsky2012imagenet},  have yielded breakthroughs in image classification and detection tasks. A large number of deep network structures and training techniques have been proposed.
		After the success of AlexNet~ \cite{krizhevsky2012imagenet}
		, deep Residual Network (ResNet)~\cite{he2016deep} has become one of the most groundbreaking networks in the deep learning community in the last few years. By using residual networks, researcher can train up to hundreds of layers and achieves excellent performance. To address the overfitting issue for training deep neural networks, the dropout technique~\cite{hinton2012improving}, ~\cite{xie2021advanced} is further proposed to regularize the model parameters by randomly dropping the hidden nodes of DNNs in the training steps to avoid co-adaptations of these nodes. Lately, the non-regularity of data structures have led to recent advancements in Graph Convolutional Networks~(GCNs)~\cite{kipf2016semi}.
		The work~\cite{hong2021graph} presents a new minibatch GCN which allows to reduce computational cost of traditional GCNs. Moreover, due to the powerful representation ability with multiple levels of abstraction, deep multi-modal representation learning~\cite{ngiam2011multimodal} has attracted much attention in recent years. The study~\cite{hong2020more}  provides a general multi-modal deep learning framework and a new fusion architecture for geoscience and remote sensing image classification applications.} 
	
	\subsection{Class Incremental Learning} 
	%\vspace{1mm}
	\zxhzxh{Continual / incremental learning~\cite{parisi2019continual} aims at learning from evolving streams of training data. 
		There are two branches of incremental learning: online incremental learning~\cite{cai2021online} that the model is single-pass through the data without task boundaries, and offline incremental learning that the model can be trained in offline mode in each incremental session. In this paper, we mainly focus on the latter one.}
	There are two major categories of incremental learning:~\emph{task incremental learning} and~\emph{class incremental learning}. A group of studies work on the task-incremental learning scenario~{\cite{PackNet,DEN,AGEM,MGR,EWC,IMM,2020Convolutional}}, where a multi-head structure is used. On the contrary, the class incremental learning~(CIL) task maintains and updates a unified classification head and thus is more challenging. \XP{This paper mainly focuses on the class-incremental learning approaches.}
	
	Class incremental learning~(CIL) is targeted at continually learning a unified classifier until all encountered classes can be recognized.
	To prevent the \emph{catastrophic forgetting} problem, a group of CIL approaches transfer the knowledge of old classes by preserving a few old classes anchor samples into the external memory buffer. Many approaches like iCaRL~\cite{2017iCaRL} and  EEIL~\cite{EEIL} use \emph{knowledge distillation} to compute the different types of distillation loss functions. {Knowledge Distillation (KD) is a technique to transfer learned knowledge from a trained neural network (as a teacher model) to a new one (as a student model)~\cite{KD,2006Model,DBLP:journals/corr/abs-2006-04719, 2018Born}. KD for class incremental learning is typically used in centralized settings before deployment in order to reduce the model complexity without weakening the predictive power~\cite{2017iCaRL,EEIL,PDR,2019Large,Hou_2019_CVPR}.}
	Later, studies such as LUCIR~\cite{2019Learning}, BiC~\cite{2019Large} and MDF~\cite{MDF} focuses on the~\emph{critical bias} problem that causes the classifier's prediction biased towards the new classes by using cosine distance classifiers or an extra bias-correction layer to fix output bias. More recently, TPCIL~\cite{tao_eccv20} puts forward the elastic Hebbian graph and the topology-preserving loss to maintain the topology of the network’s feature space. \zxhzxh{CER~\cite{JI2021107589} proposes a Coordinating Experience Replay approach to constrain the rehearsal process, which show superiority under diverse incremental learning settings.}
	Furthermore, to utilize the memory buffer more efficiently, MeCIL~\cite{9422177} proposes to keep more auxiliary low-fidelity anchor samples, rather than the original real-high-fidelity anchor samples. Nearly, \cite{9705128} recognizes the importance of the global property of the whole anchor set and designs an efficient derivable ranking algorithm to calculating loss functions.
	
	It is worth noting that existing (class) incremental learning approaches are studied in a centralized manner. They \XP{can only work in} a situation where data keeps coming from a single repository.
	Thus it cannot handle the cases where new data emerging from distributed sources.
	
	\subsection{Distributed Learning and Federated Learning}
	
	Distributed learning is mainly for training data in parallel scenarios with excessive data efficiency. Both data and workloads are divided into multiple work nodes/sites so that the burden of learning local data of each work node is within the tolerance. In each site, a local model is trained. Local models then communicate with other work nodes in accordance with certain rules like Parameter Server~\cite{Smola2010An}. The server node receives the local model from different work nodes. To integrate and build a general machine learning model, there are studies by simply averaging the model parameters to obtain a general model, solving a conformance optimization problem such as ADMM~\cite{2010distributed} and BMUF~\cite{2016Scalable}, or by model integration like ensemble learning~\cite{2017Ensemble}.
	
	Federated Learning (FL) is a popular distributed framework, which enables the creation of a general model through many local sites. The global model is aggregated by the parameters learned at the local sites on their local data. It involves training models over remote end devices, such as mobile phones.
	%There are several core challenges. Firstly, each device may have significant constraints in terms of storage, computation, and communication capacities and thus is difficult to access data across different sites. Secondly, local data that each device generates and collects is usually Non-Independent and Identically Distributed (non-IID). In addition, privacy is often a major concern in federated learning applications, and communication costs should be reduced. 
	% The typical federated learning method \emph{Federated Averaging} (FedAvg)~\cite{Mcmahan2016Communication} aggregates local parameters with the weight of the amount of data on each client, while FedMA~\cite{2020FederatedMA} proposed to match individual neurons of the neural nets layer-wise before averaging the parameters. To reduce model drifts in non-IID setting, FedProx \cite{2018Federated} incorporated a proximal term to restrict local model closer to the global model, and  Data-sharing FedAvg~\cite{DBLP:journals/corr/abs-1806-00582} used a public dataset between the server side and local sides. There are also other relevant works like FedMeta~\cite{chen2019federated}, which combined federated learning with meta learning, sharing a parameterized algorithm (or meta learner) instead of a global model. Lately, FedDF~\cite{avglogit} leverages unlabeled data to aggregate knowledge from local models to fuse a robust federated model. 
	{One typical federated learning method namely  \emph{Federated Averaging} (FedAvg)~\cite{Mcmahan2016Communication} aggregates local parameters with weights proportional to the sizes of data on each client. 
		To reduce the communication costs, STC~\cite{2019Robust} and TWAFL~\cite{0Communication} compress both the upstream and downstream communications.
		To reduce global model drifts, FedProx \cite{2018Federated} incorporates a proximal term to restrict local models closer to the global model. FedMA~\cite{2020FederatedMA} matches individual neurons of the neural networks layer-wise before averaging the parameters due to the permutation invariance of neural network parameters. There are also relevant works like FedMeta~\cite{chen2019federated}, which combine federated learning with meta learning and share a parameterized algorithm (or meta learner) instead of a global model.
		FedMAX~\cite{2020FedMAX} introduces a prior based on the principle of maximum entropy for accurate FL. 
		Moreover, data-sharing FedAvg~\cite{DBLP:journals/corr/abs-1806-00582}  uses a public dataset between the server side and local sides. Furthermore, FedDF~\cite{lin2020ensemble} leverages knowledge distillation technique to aggregate knowledge from local models to refine a robust global model, and \XP{performs} parameter averaging as done in FedAvg~\cite{Mcmahan2016Communication}. }%Particularly,
	
	Nevertheless, most of the research on distributed learning including federated learning today are still performed solving closed tasks which would hardly lead to a more open-world, long-term problem where things keep changing over time. There are only a few exceptions which is aiming at incrementally learning over multi-nodes without aggregating data~\cite{9047472,8855665}.
	However, they are in a very preliminary stage. First, they use linear neural nets, which seriously limits the applications, and fails to connect to modern deep incremental learning methods. Secondly, they only investigate a one-learning-session setting, which is apart from real continual learning over time. Thus these studies can not be applied to complicated decentralized deep incremental learning scenarios. 
	
	% \subsection{\zxh{Knowledge Distillation}}
	% \zxh{Knowledge Distillation (KD) is a technique to transfer learned knowledge from a trained neural network (as a teacher model) to a new one (as a student model)~\cite{KD,2006Model,DBLP:journals/corr/abs-2006-04719}. It is typically used in centralized settings before deployment in order to reduce the model complexity without weakening predictive power. Some works study to distill the logits of the ensemble of teacher models to a student model~\cite{2018Born}. In this work, we consider dataset \zxh{without sharing labels} for ensemble distillation, which could protect data privacy.}
	
	\section{Decentralized Class Incremental Learning}
	
	\subsection{Problem Description}
	
	We now define the Decentralized Class Incremental Learning (DCIL) problem setting as follows. The training data set $D = \{ \left ( \mathbf{x}, y \right ) \mid \mathbf{x} \in X, y \in L \}$ consists of images from an image set $X$ and their labels from a predefined common label space $L = \{1,\ldots, C\}$, where $C$ is the total number of classes. $D$ is divided into \emph{T} independent training sessions $D=\{D^{\left (1 \right )},D^{\left (2 \right )},\ldots,D^{\left (T \right )}\}$, where $D^{(t)} =  \left \{ \left ( \mathbf{x}^{(t)}, y^{(t)} \right ) \mid \mathbf{x} \in X^{(t)}, y \in L^{(t)} \right \}$. Note that the training sets of different sessions are disjoint, so do %does 
	the label sets, \emph{i.e.}, $X^{(t)} \cap X^{(p)} = \varnothing$ and $L^{(t)} \cap L^{(p)} = \varnothing$ for $t\neq p$.
	
	\XP{The goal of DCIL is to obtain a \emph{general} model $\theta^{(t)}$, which generalizes well to classify new samples of all seen classes probably appearing in any local sites.} The learning phase of each session is as follows. At session $t$, a \emph{general} model $\theta^{(t-1)}$ is prepared before the training stage of the session starts. The goal of this session is to update   $\theta^{(t-1)}$ to a \emph{new general} model $\theta^{(t)}$ so that its performance on $D^{(t)}$ can be improved. Unlike conventional incremental learning settings where the data of the session $D^{(t)}$ is centralized, in the DCIL setting, $D^{(t)}$ is decentralized and distributed to $M$ data owners  (local sites). Let $D^{(t)} = \left \{ D_1^{(t)},\cdots ,D_M^{(t)} \right \}$, where $ D_m^{(t)} = \left \{ \left ( \mathbf{x}_m^{(t)}, y_m^{(t)} \right ) \mid \mathbf{x}_m \in X_m^{(t)}, y_m \in L^{(t)} \right \}$, $X_1^{(t)}\cup X_2^{(t)}\cup \cdots \cup X_M^{(t)}= X^{(t)}$, and $X_m^{(t)} \cap X_n^{(t)} = \varnothing$ for $m \neq n$. Note that all $M$ data owners share a common class label sets of this session $L^{(t)}$. As the \emph{general} model is not allowed to access $D^{(t)}$ straightforwardly, $\theta^{(t-1)}$
	has to be distributed to each data owners and thus there are $M$ copies of $\theta^{(t-1)}$, \emph{i.e.}, $\theta_1^{(t-1)}, \cdots, \theta_M^{(t-1)}$, deployed locally to each data owners when  session learning starts. They then continually learn and update to maximize their performance on each of the $X_m^{(t)}$ separately, without forgetting the knowledge learnt in previous sessions. Finally, the learnt knowledge embedded in the updated local models $\Theta^{(t)} = \left \{ \theta_1^{(t)}, \cdots, \theta_M^{(t)} \right \}$ is transmitted to the main site for updating \emph{general} model $\theta^{(t)}$. 
	%\zxh{The goal of DCIL is to obtain a \emph{general} model $\theta^{(t)}$, which optimally generalizes to new samples from the decentralized training data in each session. }
	
	\begin{figure*}[ht]
		\begin{center}
			\includegraphics[width=0.98\textwidth]{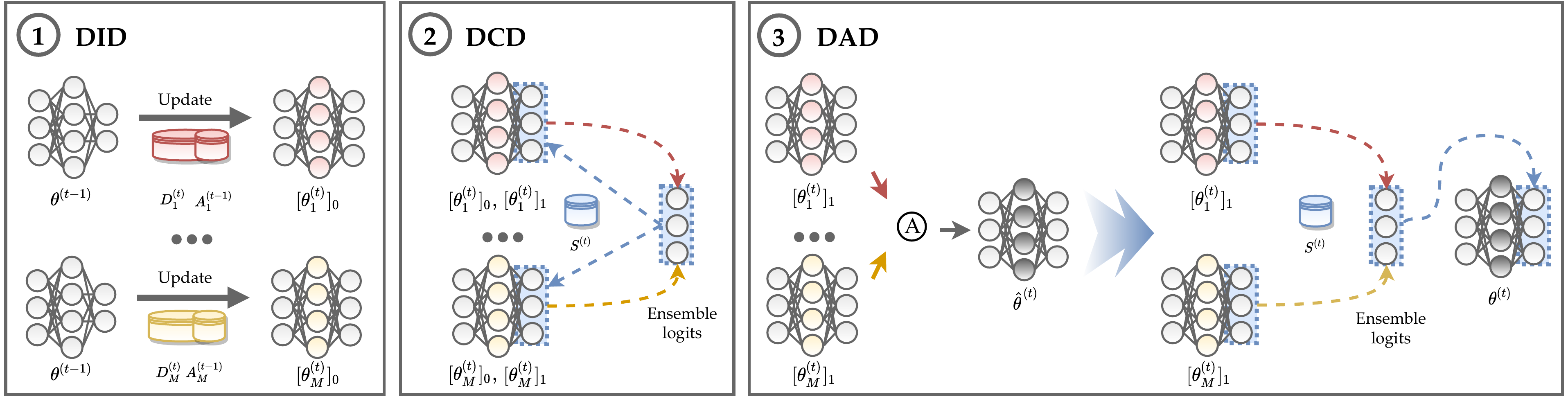}
		\end{center}
		\caption{Illustrations of the framework of DCID. The dotted line refers to the  transmission of \zxhzxh{output} on $S^{(t)}$.}
		\label{fig:exemplar}
	\end{figure*}
	
	\subsection{DCID}
	In this section, we propose the \underline{D}ecentralized \underline{C}omposite Knowledge \underline{I}ncremental \underline{D}istillation framework (DCID). As shown in Figure \ref{fig:exemplar}, DCID mainly consists of three steps. Firstly, \underline{D}ecentralized \underline{I}ncremental Knowledge \underline{D}istillation (DID) performs class incremental learning in a data decentralized setting.
	Secondly, \underline{D}ecentralized \underline{C}ollaborative Knowledge \underline{D}istillation (DCD) uses collaborated knowledge distillation among local models.
	Thirdly, \underline{D}ecentralized Knowledge \underline{A}ggregated  \underline{D}istillation (DAD) provides an aggregated knowledge distillation mechanism to update the \emph{general} model. 
	
	%In this section, we propose the \underline{D}ecentralized \underline{C}omposite Knowledge \underline{I}ncremental \underline{D}istillation framework (DCID). As shown in Figure \ref{fig:exemplar}, DCID mainly consists of three steps: \underline{D}ecentralized \underline{I}ncremental Knowledge \underline{D}istillation (DID), \underline{D}ecentralized \underline{C}ollaborative Knowledge \underline{D}istillation (DCD), and \underline{D}ecentralized Knowledge \underline{A}ggregated  \underline{D}istillation (DAD). 
	%{DID performs class incremental learning in a data decentralized setting. DCD uses collaborated knowledge distillation among local models. DAD proposes an aggregated knowledge distillation method to update the \emph{general} model.}
	% Our DCID only shares a very limited number of  \zxh{training samples without sharing their labels},
	% besides the model parameters,
	%\zxh{Our DCID only shares a very limited number of training samples without sharing their labels among local sites and the main site, besides the upload and download of model parameters,} which not only protects data privacy, but also minimizes communication cost.
	
	\vspace{2mm}
	\noindent \textbf{Decentralized Knowledge Incremental Distillation}
	\vspace{2mm}
	
	In the step of \emph{Decentralized Incremental knowledge Distillation} (DID), there are two categories of knowledge which the DID learner has to acquire: knowledge from data of the current session (\emph{i.e.}, {new-class} data) and the one from data or models of the historical sessions (\emph{i.e.}, {old-class} data). First, as constrained by the data sharing policy in the DCIL setting, new data of the session can only be accessed by corresponding local data owners (local sites) and cannot be shared by other sites. As a result, the model of the previous session has to be distributed and then deployed locally in a decentralized manner. Second, to avoid \emph{catastrophic forgetting}, it is important to transfer knowledge from models of the previous session to the local models of current session, as affirmed by \cite{8107520}, \cite{2017iCaRL}. 
	Knowledge distillation~\cite{2015Distilling} is a typical model compression and acceleration technique to transfer knowledge from large teacher models to lighter easier-to-deploy student models. This technique, as a regularizer, improves the performance of student models by giving extra soft targets with higher information entropy rather than the one-hot label (hard targets).
	Modern CIL methods usually rely on a small anchor set to maintain the memory of historical sessions, which has shown to an effective mechanism to avoid \emph{catastrophic forgetting}~\cite{2018Continual}. 
	%\dsl{Otherwise, the historical knowledge would get lost and the continuous learning process came to a halt.} 
	
	To meet these challenges, we \XP{provide a paradigm to distribute existing class} incremental learning approaches to multiple local sites where class incremental learning can be performed locally and distributively. More specifically, suppose there are $M$ local sites, the loss function $\ell_{I}^{m}$ for the $m$-th local sites in DID is composed of an anchor loss $\ell_{a}^{m}$ and a classification  loss for new data~$\ell_{c}^{m}$:
	
	\begin{equation}
	\label{eq:L_ID}
	\ell_{I}^{m} =  \ell_{c}^{m} + \zxh{\lambda}\ell_{a}^{m}, 
	\end{equation}
	\XP{where $\lambda$ is parameter controlling the strength of the two losses.}
	
	The anchor loss $\ell_{a}^{m}$ regularizes the new local model by minimizing the gap between its behavior and the old local model on the anchor set $A_{m}^{(t-1)}$ for each data owner. Through $\ell_{a}^{m}$, knowledge from models of the previous session can transfer to the local models of current session. 
	
	Generally speaking, the anchor set is composed of the most representative instances of the local data that \emph{local} models encountered in previous sessions. We use the herding method \cite{2017iCaRL,2009Herding} to construct the anchor sets for local models. For each newly encountered class $c$, the instances whose feature vector are the closest to the average feature vector \zxhzxh{$\mu_{c}^{m}$} of the class are selected as anchors, iteratively. Given the $k-1$ selected anchors, the $k$-th anchor $\mathbf{a}_{m}^{k}$ of the current class $c$ is added to the anchor set for representing this class of the \emph{local} model $[\theta_m]_0$ as follows:
	\begin{equation}
	\label{eq:getanchor}
	\mathbf{a}_{m}^{k} = \mathop{\arg\min}_{\mathbf{x}\in X_c^{m}} \left\| \zxhzxh{\mu_{c}^{m}} - \frac{1}{k}[\varphi(\mathbf{x}; [\theta_m]_0)+\sum_{j=1}^{k-1}\varphi(\mathbf{a}_m^{j};[\theta_m]_0)]\right\|,
	\end{equation}
	where $X_c^{m}$ is the instance set of class $c$ of local site $m$ and $\varphi\left ( \cdot ; \theta \right )$ is the feature function of a backbone CNN model with parameters $\theta$. Note that 
	the anchor set $A_{m}^{(t)}$ is obtained by adding new anchors from session $t$ to the previous anchor set $A_{m}^{(t-1)}$, respectively.
	
	It is worth mentioning that Eq.~\ref{eq:L_ID} provides a generic way of performing incremental learning process in local sites. By choosing a certain local incremental learner, $\ell_{a}^{m}$ and $\ell_{c}^{m}$ can be determined and $\ell_{I}^{m}$ can be minimized. Thus a large group of modern anchor based CIL approaches as mentioned in Section 2.1 can be optional. 
	\XP{In Section 4, we investigate four state-of-the-art CIL methods including iCARL~\cite{2017iCaRL}, LUCIR~\cite{Hou_2019_CVPR}, ERDIL~\cite{Dong_Hong_Tao_Chang_Wei_Gong_2021} and} TPCIL~\cite{tao_eccv20} to the DCIL setting and show that {our proposed DCID} performs well regardless of the selection of particular CIL approaches {in DID}.
	
	\vspace{2mm}
	\noindent \textbf{Decentralized Collaborated Knowledge Distillation}
	\vspace{2mm}
	
	As the local users can only access a small portion of data and are not allowed to reach the images owned by other local sites, the cluster centers estimated locally are inevitably biased. 
	%\zxh{To alleviate this bias and enhance the generalization ability, we use Knowledge Distillation (KD) technique in this work.}\zxh{Knowledge Distillation is a technique to transfer learned knowledge from a trained neural network (as a teacher model) to a new one (as a student model)~\cite{KD,2006Model,DBLP:journals/corr/abs-2006-04719}. It is typically used in centralized settings before deployment in order to reduce the model complexity without weakening predictive power. Some works study to distill the logits of the ensemble of teacher models to a student model~\cite{2018Born}.}
	We propose an efficient yet effective knowledge distillation mechanism, termed by Decentralized Collaborated Knowledge Distillation (DCD).
	
	The key idea of the DCD is making the ensemble of local models as the teacher's knowledge to guide local model learning, which has richer information entropy than the individual local model~\cite{2018large}.
	To this end, each local model (as a student), sees the ensemble of the whole local models as a teacher. Consequently, the procedure of DCD is as follows.
	
	First, at the start of the new session $t$, a small shared dataset $S^{(t)}$ is built by collecting \emph{a few samples without \XP{storing their} labels} from each local site respectively. 
	\zxh{$S^{(t)}$ contains the information that can be shared among local sites and shall be carefully set. When $|S^{(t)}|$ is too large, the burden for communicating will increase, while
		when $|S^{(t)}|$ is too small, the performance of DCD will be declined.}
	%0, our method degenerates to the basic version.
	As $S^{(t)}$ is accessible to all local sites, given an instance $\textbf{x} \in S^{(t)}$, the \zxhzxh{output (before the Softmax layer)} of each local model $[\mathbf{z}_{m}^{(t)}]_0 = f([\theta_m^{(t)}]_0, \textbf{x})$ can be computed.
	
	Second, for $\textbf{x} \in S^{(t)}$, the ensemble \zxhzxh{output} are computed by linear combination of the \zxhzxh{output} provided by local models $\{[\mathbf{z}_{m}^{(t)}]_0 = f([\theta_m^{(t)}]_0, \textbf{x})\}$. 
	We use the weighted average output as a reasonable, high-quality soft target to perform knowledge distillation. Let $[\mathbf{z}^{(t)}]_0 = \sum_{m=1}^{M}\omega_m^{(t)} [\mathbf{z}_m^{(t)}]_0$, where $\omega_m^{(t)} $ are positive numbers and $\sum_m \omega_m^{(t)} = 1$.
	
	Third, each local model learns the knowledge from the ensemble model. For each local site, the decentralized collaborated knowledge distillation loss $\ell_{C}^{m}$ over the dataset $S^{(t)}$ is minimized: 
	% \begin{equation}
	% \label{eq:L_CD}
	% \begin{aligned}
	% {\ell}_{C}^{m} = \sum_{S^{(t)}} \kappa(\frac{\exp{( [\mathbf{z}^{(t)}]_0 /\tau_{1} )}}{\sum_{j=1}^{n}\exp{([\mathbf{z}^{(t)}]_0/\tau_{1} )}}||\frac{\exp{( [\mathbf{z}^{(t)}_{m}]_0/\tau_{1} )}}{\sum_{j=1}^{n}\exp{( [\mathbf{z}_{m}^{(t)}]_0/\tau_{1} )}}),
	% \end{aligned}
	% \end{equation}

	\begin{equation}
	\label{eq:L_CD}
	\begin{aligned}
	{\ell}_{C}^{m} = \sum_{S^{(t)}} \kappa(\frac{\exp{( [\mathbf{z}^{(t)}]_0 /\tau_{1} )}}{\sum_{n}\exp{([\mathbf{z}^{(t)}]_0/\tau_{1} )}}||\frac{\exp{( [\mathbf{z}^{(t)}_{m}]_0/\tau_{1} )}}{\sum_{{n}}\exp{( [\mathbf{z}_{m}^{(t)}]_0/\tau_{1} )}}),
	\end{aligned}
	\end{equation}

	\noindent where $\tau_{1} $ is the distillation temperature parameter that controls the shape of the distribution for distilling richer knowledge from the ensemble teacher \zxhzxh{output}.
	$n=|L^{(t)}|$ is the number of new classes in the current session. The \emph{Kullback-Leibler divergence} is chosen as the fundamental distillation loss function $\kappa$.
	
	After that, each local model~$[\theta^{(t)}_m]_0$ updates to~$[\theta^{(t)}_m]_1$. It is worth mentioning that only the several samples without their labels and corresponding \zxhzxh{output} are shared between these local models. This mechanism is without any further supervised signals and it is easy to implement. Moreover, the communication cost during the process of DCD is moderate. Further, data privacy is guaranteed as only \emph{a few images} are shared, without providing corresponding labels.
	
	\vspace{2mm}
	\noindent \textbf{Decentralized Aggregated Knowledge Distillation}
	\vspace{2mm}
	
	In the final stage, the Decentralized Aggregated knowledge Distillation method transfers the knowledge of multiple local models back to the main site to update the general model~$\theta^{(t)}$. One naïve solution is to use the most popular aggregation method in Federated learning, i.e., FedAvg~\cite{Mcmahan2016Communication} to fulfill this function. Nevertheless, the shared dataset $S^{(t)}$ and the set of ensemble \zxhzxh{output} maintained by the second stage DCD provides extra opportunities to distill knowledge and further improve the aggregation results. As a result, {we design an effective two-step DAD solution  to refine the general model} as follows:
	
	At first, the general model $\hat{\theta}^{(t)}$ is initialized by aggregating all local models using FedAvg~\cite{Mcmahan2016Communication}. That is, general model $\hat{\theta}^{(t)}$ aggregates local models $\left \{ [\theta_1^{(t)}]_1, \cdots, [\theta_M^{(t)}]_1 \right \}$ by weighted averaging:
	\begin{equation}
	\label{eq:fedavg}
	{\hat{\theta}}^{(t)} = \sum_{m=1}^{M}\frac{N_m^{(t)}}{N^{(t)}}[\theta_m^{(t)}]_1,
	\end{equation}
	where $N^{(t)}$ is the number of training samples over all local sites, $N_m^{(t)}$ is the number of training samples in the $m$-th local site.
	
	Then, the main site distills the ensemble of all the local models (as teachers) to one single general model (as a student). To this end, we use the same ensemble approach as DCD process. The local models are evaluated on mini-batches of data from the current shared dataset $S^{(t)}$ and their \zxhzxh{output} are aggregated as teacher \zxhzxh{output} $[\mathbf{z}^{(t)}]_1$. The student \zxhzxh{output} $[\hat{\mathbf{z}}^{(t)}]_1 = f([\hat{\theta}_m^{(t)}], \textbf{x})$  are generated from the inital general model evaluated on $S^{(t)}$.
	
	To secure an enhanced performance, data on $S^{(t)}$ and the corresponding teacher \zxhzxh{output} and student \zxhzxh{output} are shuffled. We then minimize the decentralized aggregated knowledge distillation loss $\ell_{A}$ over the dataset $S^{(t)}$:
	% \begin{equation}
	% \begin{aligned}
	% \label{eq:AKD}
	% % L_{FedAD} =  \sum_{i}^{N_a}\tau^2_{2}\mathbf{p}^{\tau_{2}}_{i}\log{\frac{\mathbf{p}^{\tau_{2}}_{i}}{\mathbf{q}^{\tau_{2}}_{i}}}.
	% \ell_{A} = \sum_{S^{(t)}} \kappa(\frac{\exp{( [\mathbf{z}^{(t)}]_1 /\tau_{2} )}}{\sum_{j=1}^{n}\exp{([\mathbf{z}^{(t)}]_1/\tau_{2} )}}||\frac{\exp{( [\hat{\mathbf{z}}^{(t)}]_1/\tau_{2} )}}{\sum_{j=1}^{n}\exp{( [\hat{\mathbf{z}}^{(t)}]_1/\tau_{2} )}}).
	% \end{aligned}
	% \end{equation}

	\begin{equation}
	\begin{aligned}
	\label{eq:AKD}
	\ell_{A} = \sum_{S^{(t)}} \kappa(\frac{\exp{( [\mathbf{z}^{(t)}]_1 /\tau_{2} )}}{\sum_{{n}}\exp{([\mathbf{z}^{(t)}]_1/\tau_{2} )}}||\frac{\exp{( [\hat{\mathbf{z}}^{(t)}]_1/\tau_{2} )}}{\sum_{{n}}\exp{( [\hat{\mathbf{z}}^{(t)}]_1/\tau_{2} )}})
	.
	\end{aligned}
	\end{equation}

	\noindent We use the same form of $\kappa$ as in Eq.~\ref{eq:L_CD}. $\tau_{2} $ is the distillation temperature parameter. Note that we only transfer corresponding \zxhzxh{output} of $S^{(t)}$ from each local site to the main site, with only a small amount of extra communication costs as the same as DCD. 
	
	\vspace{2mm}
	\noindent \textbf{Overall Learning Procedure}
	\vspace{2mm}
	
	We recap the operations performed by the local site and the main site at session $t$  as follows.
	
	Each \textbf{local site} uploads the random chosen %unlabeled samples 
	\zxh{samples without sharing their labels}
	of new classes to the main site, and downloads the shared dataset $S^{(t)}$. Then the following processes are repeated until convergence. Firstly, each local site updates the general model $\theta^{(t-1)}$ of the previous session using Eq.~\ref{eq:L_ID} to obtain $[\theta_m^{(t)}]_0$.
	Secondly, each local site performs DCD, finetunes its model using the ensemble \zxhzxh{output} according to Eq.~\ref{eq:L_CD} and obtains $[\theta_m^{(t)}]_1$. Thirdly, each local site uploads its model parameter $[\theta_m^{(t)}]_1$ and \zxhzxh{output} on $S^{(t)}$ to the main site.
	
	The \textbf{main site} receives the uploaded samples from all local sites and construct a shared dataset $S^{(t)}$. Then the following processes are repeated for $R$ \emph{round}s.
	Each \emph{round} represents the process that the general model to be distributed, and the new general model is obtained after local model training and aggregation.
	The main site firstly broadcasts the latest general model $\theta^{(t-1)}$ to local sites. Secondly, it calculates the ensemble \zxhzxh{output} by using the uploaded \zxhzxh{output} from local sites and distributes them back to the local sites. Thirdly, the main site aggregates model parameters sent from all the local sites by taking an weighted average of them to initialize the general model parameters $\hat{\theta}^{(t)}$, then finetunes it by using the ensemble of \zxhzxh{output} from local sites to get $\theta^{(t)}$. Finally, the main site broadcasts the updated $\theta^{(t)}$ to all local sites.
	
	% \begin{figure*}[thbp!]
	%  \includegraphics[width=1\textwidth]{figures/compareResult_2R3C.pdf}
	% \caption{Comparison between our method and baseline on CIFAR100 and subImageNet. The symbols i, L, and T in the parenthesis are short for iCARL, LUCIR, and TPCIL, respectively. All solid curves stand for our \emph{DCID}, while the dotted curves for the baseline. The accuracy values reported here are the mean of five-times test.}
	% \label{fig:sub}
	% \end{figure*}
	
	\begin{figure*}[thbp!]
		\begin{center}
			\includegraphics[width=1\textwidth]{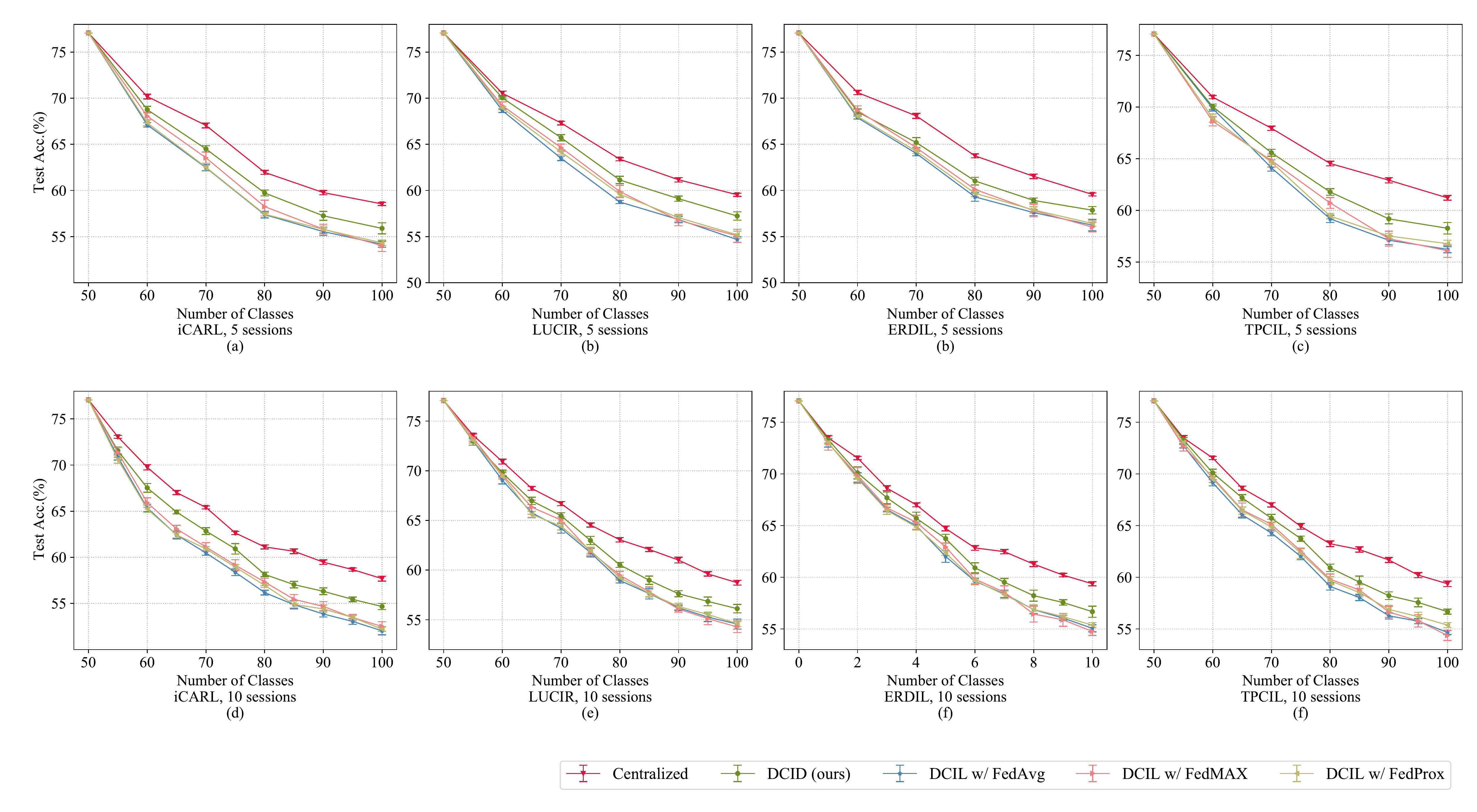}
		\end{center}
		\caption{Comparison between DCID and baselines on CIFAR100. The accuracy values reported here are the mean and the stand deviation of three-times test.}
		\label{fig:cifar}
	\end{figure*}

	The overall learning procedure of the proposed decentralized composite knowledge incremental distillation framework in one session is summarized in Algorithm~\ref{algo:dcil}. \XP{It is worth mentioning that besides the upload and download of model parameters, our DCID only shares a very limited number of training samples without sharing their labels among local sites and the main site,} which not only protects data privacy, but also minimizes the communication cost.
	
	%\zxh{Our DCID only shares a very limited number of training samples without sharing their labels among local sites and the main site, besides the upload and download of model parameters,} which not only protects data privacy, but also minimizes communication cost.

	\subsection{Baseline Approaches}%{A basic DCIL framework} %distributing framework}
	\label{sec:generic_framework}
	
	As we focus on a new learning paradigm, it is desired to provide baseline results and build a benchmark for the decentralized class incremental learning study. \XP{To meet this need, we develop a basic decentralized framework to expand typical class-incremental learning methods like the \emph{four} introduced in the previous sub-section to their DCIL counterparts.} {The overall procedure of the proposed basic DCIL framework are \XP{summarized in Algorithm~\ref{algo:baseline}} and described as follows:}
	%federated learning
	%so that a group of baseline methods can be developed and provide 
	%a baseline DCIL method for evaluations.
	%\zxh{baseline approaches for evaluations.}
	
	%\XP{To provide baseline DCIL approaches, we develop a framework, where typical existing deep class-incremental learning methods and  using some state-of-the-art federated learning methods in DCIL problem setting.} 
\begin{algorithm}
\caption{The DCID framework}\label{algo:dcil}
  \SetKwData{Left}{left}\SetKwData{This}{this}\SetKwData{Up}{up}
  \SetKwFunction{Union}{Union}\SetKwFunction{FindCompress}{FindCompress}	
  \SetKwInOut{Input}{Input}\SetKwInOut{Output}{Output}
   \Input{Training set $\{ D_1^{(t)},\cdots ,D_M^{(t)}  \}$ of the current session, the anchor set $ \{ A_1^{(t-1)},\cdots ,A_M^{(t-1)} \}$, the shared dataset $S^{(t)}$ and the general model $\theta^{(t-1)}$ after the previous session. }
  \Output{The updated general model $\theta^{(t)}$ of the current session.}
  Copy the general model $\theta^{(t-1)}$ of the main site and distribute to $M$ data owners as $\left \{ \theta^{(t)}_{m} \right \}$\;
  \For{\rm{each round} $r=0, 1,...,$ \KwTo $R-1$}{
    \For{\rm{each local site} $m=0,1,...,$ \KwTo $M-1$ \rm{in parallel}}{
     Update local models $[\theta^{(t)}_{m}]_0$ by minimizing $\ell_{I}^{m}$\;
      Compute local \zxhzxh{output} $[z^{(t)}_m]_0$ on $S^{(t)}$\;
      Compute~$A^{(t)}_m$~according to Eq.\ref{eq:getanchor}\;
    }
    $[z^{(t)}]_0 = \sum_{m=1}^{M}\omega_{m}^{(t)} [z^{(t)}_m]_0$\;
    \For{\rm{each local site} $m=0,1,...,$ \KwTo $M-1$ \rm{in parallel}}{
     Update local models $[\theta^{(t)}_{m}]_1$ by minimizing $\ell_{C}^{m}$\;
    }
    $\hat{\theta}^{(t)} \leftarrow \sum_{m=1}^{M}\frac{N_m}{N}[\theta^{(t)}_{m}]_1$\;
    \For{\rm{each local site} $m=0,1,...,$ \KwTo $M-1$ \rm{in parallel}}{
    Compute local \zxhzxh{output} $[z^{(t)}_m]_1$ on $S^{(t)}$\;
    }
    $[z^{(t)}]_1 = \sum_{m=1}^{M}\omega_{m}^{(t)} [z^{(t)}_m]_1$\;
    Update general model $\theta^{(t)}$ by minimizing $\ell_A$\;
  }
\end{algorithm}	
	
\begin{algorithm}
\caption{The basic DCIL framework}\label{algo:baseline}
  \SetKwData{Left}{left}\SetKwData{This}{this}\SetKwData{Up}{up}
  \SetKwFunction{Union}{Union}\SetKwFunction{FindCompress}{FindCompress}
  \SetKwInOut{Input}{Input}\SetKwInOut{Output}{Output}
   \Input{Training set $\{ D_1^{(t)},\cdots ,D_M^{(t)}  \}$ of the current session,  the anchor set $ \{ A_1^{(t-1)},\cdots ,A_M^{(t-1)} \}$,
  and the general model $\theta^{(t-1)}$ after the previous session.}
  \Output{The updated \emph{general} model $\theta^{(t)}$ of the current session.}
  Copy the general model $\theta^{(t-1)}$ of the main site and distribute to $M$ data owners as $\left \{ \theta^{(t)}_{m} \right \}$\;
  \For{\rm{each round} $r=0, 1,...,$ \KwTo $R-1$}{
    \For{\rm{each local site} $m=0,1,...,$ \KwTo $M-1$ \rm{in parallel}}{
      Update local models $\theta^{(t)}_{m}$ by minimizing ${\ell}_{loc}$\;
      Compute~$A^{(t)}_m$~according to Eq.~\ref{eq:getanchor}\;
    }
    $\theta^{(t)} \leftarrow \sum_{m=1}^{M}\frac{N_m}{N}\theta^{(t)}_{m}$\;
  }
\end{algorithm}
	
	Given a particular class-incremental learner, at the beginning of session $t$, the main site distributes the general model $\theta^{(t-1)}$ to $M$ local data owners (\emph{Step 1}). Then, each of the data owners updates the local model using its own training data $X^{(t)}_{m}$ to adapt to new classes while performing a certain kind of forgetting alleviation mechanism using a few old-class anchors $A_m^{(t-1)}$ (\emph{Step 4}). Note that $X^{(t)}_{m}$ and $A_m^{(t-1)}$ in each local site will not be transmitted to other local sites or the main site. \XP{The local anchor set $A_m^{(t-1)}$ is then updated to be $A_m^{(t)}$ (\emph{Step 5}). Finally, the updated local models are transmitted to the main site and the 
		general model evolves as the simple weighted average of all the local models  based on the amount of data trained at the current session (\emph{Step 7}).} \XP{It is worth mentioning that Algorithm~\ref{algo:baseline} works as well for incremental learning methods without using a historical anchor set, by simply setting $A = \varnothing$ and removing Step 5.}
	
	%the most popular aggregation method ~\cite{Mcmahan2016Communication} (\emph{Step 7}). More specifically, the \emph{general} model is computed as the weighted average of all the local models  based on the amount of data trained at the current session using \zxh{Eq.}~\ref{eq:fedavg}. 
	
	\XP{The proposed framework is carefully designed so that popular federated updating and aggregation methods such as FedAvg, FedMAX~\cite{2020FedMAX}, and FedProx~\cite{2018Federated} can be used as a plug-in module to update the local models and gather them as a whole into a general one. The selection of these methods can be easily controlled by defining different \emph{${\ell}_{loc}$} in \emph{Step 4}. When using FedAvg for aggregation, \emph{${\ell}_{loc}$} is defined as follows and the resulted DCID method is referred to as ``DCIL w/ FedAvg":}
	\begin{equation}
	\label{eq:dcid_fedavg}
	{\ell}_{loc} = {\ell}_{c}^{m} + \lambda{\ell}_{a}^{m},
	\end{equation}
	where ${\ell}_{c}^{m}$, $\lambda$, and ${\ell}_{a}^{m}$ are the same with those in Eq.~\ref{eq:L_ID}. 
	Alternatively, the methods ``DCIL w/ FedMAX" can be defined as: 
	\begin{equation}
	\label{eq:dcid_fedmax}
	{\ell}_{loc} = {\ell}_{c}^{m} + \lambda{\ell}_{a}^{m} + \beta \frac{1}{B}\sum_{i=1}^{B}\kappa(\zxh{a_i^m}\|U),
	\end{equation}
	{where $a_i^m$ refers to the activation vector at the input of the last fully-connected layer for sample $i$ on the $m$-th local site. $U$ stands for a uniform distribution over the activation vectors. $B$ is a mini-batch size of local training data. $\kappa$ is the \emph{Kullback-Leibler divergence} and $\beta$ is a hyper-parameter used to control the scale of the regulation loss.}
	Moreover, ${\ell}_{loc}$ for ``DCIL w/ FedProx" becomes
	\begin{equation}
	\label{eq:dcid_fedprox}
	{\ell}_{loc} = {\ell}_{c}^{m} + \lambda{\ell}_{a}^{m} + \frac{\mu}{2}\| \theta_m^{(t)} - \theta^{(t)}\|,
	\end{equation}
	where $\mu$ is the parameter controlling the scale of the proximal term. Please refer to~\cite{2020FedMAX} and~\cite{2018Federated} for more details of the FedMAX and FedProx methods.

	%\zxh{We also apply FedMAX~\cite{2020FedMAX} and FedProx~\cite{2018Federated} methods to DCIL setting, i.e., ``DCIL w/ FedMAX" and ``DCIL w/ FedProx", which can be described in Algorithm~\ref{algo:baseline}. FedMAX and FedProx are extension methods of FedAvg. FedMAX proposes a prior $U$ based on the principle of maximum entropy to enable similar activation vectors $a_m^i$ across multiple data owners to address the activation-divergence issue. FedProx incorporates a proximal term for the local training to better accuracy. We use \emph{${\ell}_{fed}$} to represent their role in local model updating. For FedAvg, ${\ell}_{fed} = {\ell}_{c}^{m} + \lambda{\ell}_{a}^{m}$. For FedMAX, ${\ell}_{fed} = {\ell}_{c}^{m} + \lambda{\ell}_{a}^{m} + \beta \frac{1}{L^{(t)}}\sum_{i=1}^{L^{(t)}}\tau(a_m^i\|U)$; for FedProx, ${\ell}_{fed} = {\ell}_{c}^{m} + \lambda{\ell}_{a}^{m} + \frac{\mu}{2}\| \theta_m^{(t)} - \theta^{(t)}\|$.} 
	
	\section{Experiments}
	\label{sec:expr}
	To facilitate the study of DCIL, we conduct comprehensive experiments under the DCIL setting to provide baseline results of the DCIL frameworks and evaluate the proposed DCID approach.
	%In order to verify the effectiveness of our method, we conduct comprehensive experiments under the DCIL setting on two popular image classification datasets CIFAR100 \cite{Krizhevsky2009Learning} and subImageNet \cite{2017iCaRL}\cite{2019Learning}.
	
	\subsection{Data and Setup}
	
	Experiments are performed in PyTorch under the DCIL setting on two challenging image classification datasets CIFAR100 \cite{Krizhevsky2009Learning} and subImageNet \cite{2017iCaRL,2019Learning}. %\XP{All experiments are performed in PyTorch.}
	
	\textbf{CIFAR100} contains 60,000 natural RGB images over 100 classes, including 50,000 training images and 10,000 test images. It is very popular both in incremental learning works~\cite{2018End,2017iCaRL} and distributed / federated learning~\cite{Mcmahan2016Communication} works. Each image has a size of 32×32. We randomly flip images for data augmentation during training.
	
	\textbf{SubImageNet} contains images of 100 classes randomly selected from ImageNet\cite{Russakovsky2015ImageNet}. There are about 130,000 RGB images for training and 5,000 RGB images for testing. For data augmentation during training, we randomly flip the image and crop a 224×224 patch as the network’s input. During testing, we crop a single center patch of each image for evaluation.
	
	\vspace{1mm}
	\noindent \textbf{Experimental setting}
	\vspace{1mm}

	For each dataset, we randomly choose half of the classes, \emph{i.e.}, 50 classes as the base classes for the base session (\emph{session 0}) and divide the rest of the classes into~$T=5$ and $10$ sessions for incremental learning.
	
	In each session, there are $M$ local sites where the \emph{general} model of the previous session is copied, deployed, and updated locally. We randomly sample the dataset of each session under a balanced and independent and identical distribution (IID) \zxh{for experiments in Sections \ref{subses:Comparison Results}, \ref{subses:Ablation Study} and \ref{subses:key issues}}. Evaluations for non-IID settings are provided in Section~\ref{subses:Heterogeneous}. $M$ is set to 5 as a basic setting for most experiments in Sections {\ref{subses:Comparison Results}, \ref{subses:Ablation Study}}, \ref{subses:key issues} and \ref{subses:Heterogeneous}.
	More comprehensive evaluations about different settings of $M$ in a range of $\{2, 5, 10, 20\}$ are provided as well in Section {\ref{subses:key issues}}.%4.4.
	
	%\zxh{Data partitioning is adopted for generating decentralized datasets by sampling the dataset of each session under balanced and independent and identical distribution (I.I.D.). The training data of each session is shuffled, and then partitioned into each local sites, similar to McMahan et al.~\cite{Mcmahan2016Communication}.}
	
	\begin{figure*}[ht]
		\begin{center}
			\includegraphics[width=1\textwidth]{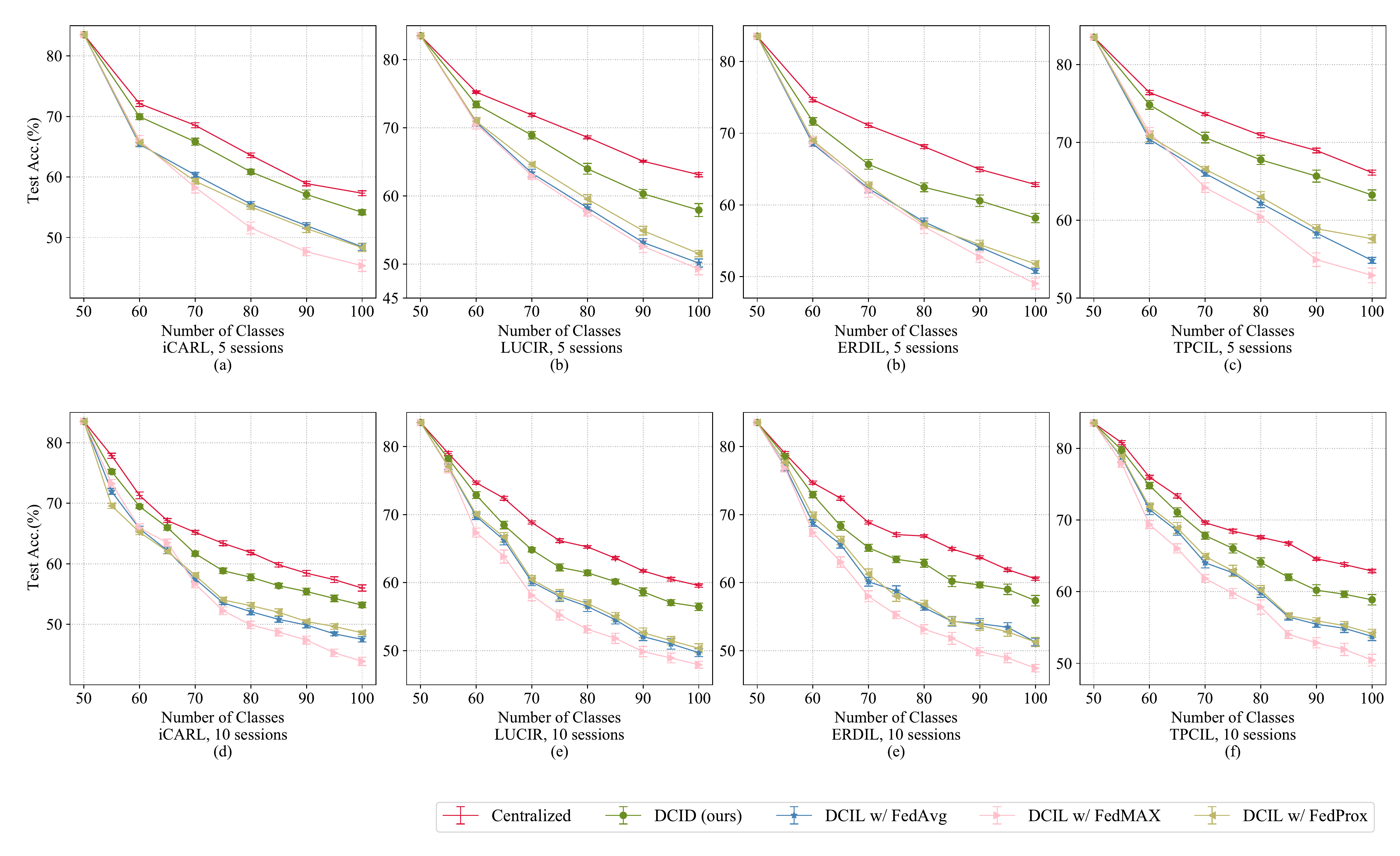}
		\end{center}
		\caption{Comparison between DCID and baselines on subImageNet. The accuracy values reported here are the mean \zxh{and the stand deviation} of three-times test.}
		\label{fig:sub}
	\end{figure*}
	
	We choose \emph{four} representative class incremental learning approaches: \textbf{iCARL}~\cite{2017iCaRL}, \textbf{LUCIR}~\cite{Hou_2019_CVPR}, \textbf{ERDIL}~\cite{Dong_Hong_Tao_Chang_Wei_Gong_2021} and \textbf{TPCIL}~\cite{tao_eccv20} as the basic incremental learners. iCARL classifies by using the nearest-mean-of-exemplars rule and the~\emph{traditional knowledge distillation} to transfer the knowledge. LUCIR incorporates three components, i.e., cosine normalization, \emph{feature knowledge distillation}, and inter-class separation, to alleviate catastrophic forgetting.
	{ERDIL uses an exemplar relation graph to explore the relations information of exemplars from the old classes and leverages the \emph{graph-based relational knowledge distillation} to transfer old knowledge for new class learning.}
	TPCIL maintains the topology knowledge space by elastic Hebbian graph and \emph{typology-preserving loss}. We decentralize them to multiple local sites in the DID stage to verify the proposed framework.
	At the start of each session, each local site initializes classification layer parameters for new classes equally before training for better convergence~\cite{Mcmahan2016Communication}.
	
	A series of session accuracy are recorded on the testing sets at the end of each session from \emph{session 0} to the last session, respectively. Two \textbf{measures}, \emph{i.e.}, the \emph{average accuracy} over such a series of {accuracy} as well as the \emph{final accuracy} namely the accuracy of the last session are reported for evaluating the performance~\cite{2020Mnemonics,2017iCaRL,tao_eccv20,tao2020few,Hou_2019_CVPR,Dong_Hong_Tao_Chang_Wei_Gong_2021}. 
	The details settings of the parameters are reported as follows. 
	
	The anchor number is set to 20 for each class of local sites and the number of shared data for each class is set to 20. During knowledge distillation, the learning rate is set to $10^{-4}$ and the epoch is set to 5. The distillation temperature parameter in Eqs.~\ref{eq:L_CD} and \ref{eq:AKD} are set to 5. 
	\XP{We have evaluated $\mu$ in Eq.~\ref{eq:dcid_fedprox} for ``DCIL w/ FedProx" in the interval of $[0.02, 10]$ and found that the results are insensitive to the setting. As a result, we set $\mu = 0.2$. Analogously,}
	we set $\beta$ = 500 in Eq.~\ref{eq:dcid_fedmax} for ``DCIL w/ FedMAX".
	%\zxh{Further, we use $\mu = 0.2$ in Eq.~\ref{eq:dcid_fedprox} for ``DCIL w/ FedProx". We did try other $\mu$ values like $\{0.02, 0.5, 1, 10\}$, but found that the results are very similar. }
	
	Considering different scales of the two datasets, we follow mainstream class-incremental learning study to choose different backbone networks with corresponding settings.
	On CIFAR100, we choose the popular 32-layer ResNet~\cite{he2016deep} as the backbone, as in ~\cite{2019Learning}. Initially, we train the base model for 160 epochs using minibatch SGD with a minibatch size of 128. As per the recommended settings from the original papers, we set the hyper-parameter \zxh{$\lambda=15$} in Eq.\ref{eq:L_ID} for TPCIL, $\lambda=100$ for ERDIL while \zxh{$\lambda=5$} for iCARL and LUCIR. The initial learning rate is set to 0.1 and decreased to 0.01 and 0.001 at epoch 80 and 120, respectively.  During decentralized deep incremental learning sessions, we choose the local epoch $E$ = 10, with a minibatch size of 128 for each local site. The learning rate is initially set to 0.01 and decreased by 10 times at epoch 10. The number of rounds~$R$ is set to 10 in both Algorithms~\ref{algo:dcil} \XP{and~\ref{algo:baseline}}.
	On subImageNet, we follow~\cite{Hou_2019_CVPR} and use the 18-layer ResNet as the backbone.  We set the hyper-parameter \zxh{$\lambda=100$} in Eq.~\ref{eq:L_ID} for ERDIL and $\lambda=10$ for other CIL methods. We train the base model with a minibatch size of 128 and the initial learning rate is set to 0.1. We decrease the learning rate to 0.01 and 0.001 after epoch 30 and 50, respectively, and stop training at epoch 100.  Then, we fintune the model on each subsequent and decentralized training set. The learning rate is initially set to 0.1 and decreased by 10 times at epoch 10. We choose the local epoch $E$ = 30, with a minibatch size of 128 for each local site. $R$ is set to 3.
	
	% \begin{figure*}[thbp!]
	% \begin{center}
	%     \includegraphics[width=1\textwidth]{figures/qikanResult_sub.pdf}
	% \end{center}
	% \caption{Comparison between our method and baselines on subImageNet. The accuracy values reported here are the mean of three-times test.}
	% \label{fig:sub}
	% \end{figure*}
	
	\subsection{Comparison Results \XP{under the IID setting}}
	\label{subses:Comparison Results}
	\zxh{
		% Traning data in session $t$ follows a distribution over $L^{(t)}$ classes parameterized by a vector $\bf q$ ($q_i \geq 0, i \in [1, L^{(t)})$ and $||\bf q||_1=1$).
		Under the IID setting, the training data of each new class in session $t$ is partitioned into each local sites randomly, similar to McMahan~\cite{Mcmahan2016Communication} et al.} 
	Figure~\ref{fig:cifar} and~\ref{fig:sub} show the comparison results between our proposed DCID and baseline approaches described in Section~\ref{sec:generic_framework} on two datasets: CIFAR100 and subImageNet. For stable evaluation results, we run the experiments for {\textbf{three} times} and report the mean results using four representative class-incremental methods: iCARL, LUCIR, ERDIL and TPCIL as reported. Each curve reports the mean testing accuracy over sessions.
	\XP{To achieve upper-bound performances for reference,} {we retrain the model at each session with {a centralized setting of data}, which is denoted as ``Centralized".}
	%with the green, blue and pink colors standing for the tests using iCARL, LUCIR, and TPCIL, respectively.
	{The green curve reports the accuracy achieved by our proposed DCID, while the crimson, blue, pink, and amber curves report the accuracies of ``Centralized", ``DCIL w/ FedAvg", ``DCIL w/ FedMAX", and ``DCIL w/ FedProx", respectively.}
	The main results are summarized as follows:

	\begin{itemize}
		\item 
		% In all six experiments with all three session settings on both two datasets, our DCID consistently outperforms the baseline method on each incremental session by a large margin, especially on the challenging subImageNet. The superiority of our method becomes more obvious after learning all the sessions, which shows the effectiveness of long-term incremental learning with decentralized data.
		In all experiments with 5 and 10 session settings on both two datasets, our DCID consistently outperforms baseline results on each incremental session by a large margin, especially on the challenging subImageNet. The superiority of our method becomes more obvious after learning all the sessions, which shows the effectiveness of long-term incremental learning \XP{from} decentralized data.
		
		\item On CIFAR100, our DCID frameworks with iCARL, LUCIR, ERDIL and TPCIL achieve average accuracies of 63.87\%, 65.05\% ,64.76\%, and 65.31\% respectively under the 5-session setting. In comparison, the second-best ``DCIL w/ FedMAX" frameworks achieves the average accuracies of 62.78\%, 63.77\%, 64.05\% and 64.86\%, respectively. 
		As a result, DCID outperforms the second-best one by up to 1.28\% \zxhzxh{in term of the average accuracy with LUCIR method}. After learning all the sessions, DCID further outperforms ``DCIL w/ FedMAX" by up to 1.88\% \zxhzxh{in term of the final accuracy with LUCIR method}. 
		Analogously, under the 10-session setting, our DCID also outperforms baseline approaches both in average accuracy and final accuracy.
		
		\item {On subImageNet, under the 5-session setting, our DCID with iCARL, LUCIR, ERDIL and TPCIL achieves the average accuracies of 65.23\%, 68.01\%, 66.99\% and 70.93\%, respectively, which are 4.36\%, 3.83\%, 3.88\% and 4.20\% higher than the second best method, correspondingly. Furthermore, at the last session, our DICD greatly outperforms the second best method by up to 5.70\%, 6.36\%, 6.40\% and 5.62\%. 
			Under the 10-session setting, our DCID also outperforms the second best method with iCARL, LUCIR, ERDIL and TPCIL by 4.35\%, 3.77\%, 4.14\% and 3.15\% in terms of average accuracies, respectively. 
			Moreover, at the end of the entire learning process, our method outperforms the second best method  by 4.59\%, 6.10\%, 6.08\% and 4.69\%, respectively.}
	\end{itemize}
	
	\subsection{Ablation Study}
	\label{subses:Ablation Study}
	\vspace{2mm}
	\zxhzxh{We provide ablation studies on subImageNet to investigate each component’s contribution to the final performance gain and prove the effectiveness and generalization ability of DCID.}
	\begin{table}[h!]
		\centering  
		\caption{Abation study on subImageNet with iCARL.}
		\label{ic}
		\renewcommand\arraystretch{1.2}
		\renewcommand\tabcolsep{2.5pt}
		\footnotesize
		\begin{tabular}{lccccccc}
			\bottomrule
			\multirow{2}{*}{Component}  & \multicolumn{6}{c}{Encountered Classes} & \multicolumn{1}{c}{Average} \\ \cline{2-7}
			& 50 & 60 & 70 & 80 & 90 & 100 & \multicolumn{1}{c}{Acc.} \\ \hline\hline
			Baseline & 83.52 &65.39 & 60.34 & 55.52 &51.97 & 48.46 & 60.86\\
			Baseline w/ DCD  & 83.52 & 67.28&	64.67&	59.24&	55.98&	53.79&	64.05   \\
			Baseline w/ DAD  & 83.52 & 67.21&	63.93&	58.62&	54.71&	53.67&	63.61  \\
			DCID & 83.52 & \textbf{69.96}&\textbf{65.82}&	\textbf{60.85}&	\textbf{57.11}&	\textbf{54.16}&	\textbf{65.23}  \\ \bottomrule
		\end{tabular}
	\end{table}
	
	\begin{table}[h!]
		\centering  
		\caption{Abation study on subImageNet with LUCIR.}
		\label{lu}
		\renewcommand\arraystretch{1.2}
		\renewcommand\tabcolsep{2.5pt}
		\footnotesize
		\begin{tabular}{lccccccc}
			\hline\hline
			\multirow{2}{*}{Component} & \multicolumn{6}{c}{Encountered classes} & \multicolumn{1}{c}{Average} \\ \cline{2-7}
			& 50 & 60 & 70 & 80 & 90 & 100 & \multicolumn{1}{c}{Acc.} \\ \hline
			Baseline& 83.52 & 70.83& 63.34& 58.25&53.16&50.14&63.21 \\
			Baseline w/ DCD  & 83.52 & 71.93&	68.14&	63.7&	59.51&	56.48&	67.21  \\
			Baseline w/ DAD  & 83.52 & 71.64&	67.02&	62.02&	57.88&	55.72&	66.30   \\
			DCID & 83.52 & \textbf{73.43}& 	\textbf{68.91} &	\textbf{63.98}&\textbf{60.31} &	\textbf{57.92}&	\textbf{68.01}   \\ \hline\hline
		\end{tabular}
	\end{table}
	
	\begin{table}[h!]
		\centering  
		\caption{Abation study on subImageNet with TPCIL.}
		\label{tp}
		\renewcommand\arraystretch{1.2}
		\renewcommand\tabcolsep{2.5pt}
		\footnotesize
		\begin{tabular}{lccccccc}
			\hline\hline
			\multirow{2}{*}{Component}  & \multicolumn{6}{c}{Encountered Classes} & \multicolumn{1}{c}{Average} \\ \cline{2-7}
			& 50 & 60 & 70 & 80 & 90 & 100 & \multicolumn{1}{c}{Acc.} \\ \hline
			Baseline & 83.52&	70.39&66.01&62.2&58.35&54.84&65.89 \\
			Baseline w/ DCD & 83.52&73.68&	69.26&	66.43&	62.27&	61.17&	69.38  \\
			Baseline w/ DAD & 83.52&	72.92&	68.01&	65.28&	61.14&	59.73&	68.43   \\
			DCID & 83.52 &	\textbf{74.83} & \textbf{70.62}&\textbf{67.75}&\textbf{65.67}&\textbf{63.24}&\textbf{70.93} \\ \hline\hline
		\end{tabular}
	\end{table}
	
	\zxhzxh{We conduct the experiments using three different centralized CIL approaches: iCARL, LUCIR and TPCIL. The experiments are performed on subImageNet under the 5-session incremental learning setting with 5 local sites. We explore the impact of the Decentralized Collaborate knowledge Distillation (DCD) module in Eq.\ref{eq:L_CD} and Decentralized Aggerated knowledge Distillation (DAD) module in Eq.\ref{eq:AKD}, respectively. 
		Table~\ref{ic}, \ref{lu}, and 
		\ref{tp} report the comparative results using iCARL, LUCIR and TPCIL methods, respectively.
		Our ``Baseline" method is “DCIL w/ FedAvg”, in other words, removing the DAD and DCD modules from DCID.
		%  Recall that ``DCID" is ``Baseline" 
		% with both DAD and DCD. 
		``Baseline w/ DAD" refers to ``Baseline" method with the DAD module. We summarize the results as follows:}
	
	\begin{itemize}
		\item  \zxhzxh{ 
			Both the DCD and the DAD modules improve the performance of the baseline no matter which of the three basic CIL methods is used. \emph{Baseline with DCD} outperforms the baseline by up to 3.19\%, 4.00\%, and 3.49\% using iCARL, LUCIR and TPCIL, respectively. \emph{Baseline with DAD} exceeds the baseline by up to 2.75\%, 3.09\%, and 2.54\%, correspondingly.}
		\item  \zxhzxh{When combing the two modules with the baseline, DCID achieves the best average accuracy. It exceeds \emph{Baseline with DCD}  by up to 1.18\%, 0.80\% and 1.55\% using iCARL, LUCIR and TPCIL, and \emph{Baseline with DAD} by up to 1.64\%, 1.71\% and 2.50\%, respectively.}
	\end{itemize}
	
	\zxhzxh{All these results show that our proposed method consistently outperforms the baselines, regardless of the (centralized) CIL methods used. The effectiveness of the proposed DCID and its components is demonstrated.}
	
	\subsection{Key Issues of DCID}
	\label{subses:key issues}
	
	\vspace{2mm}
	\noindent \textbf{The effect of the size of shared dataset}
	\vspace{2mm}
	
	To investigate the effect brought by the different amounts of the shared dataset, we further evaluate the methods using the different numbers of shared samples per class on a local site. With the 5 session and 5 local site settings, we compare the performance comparison using LUCIR method in Table~\ref{ss}.  The number of shared samples per class is in the range of $\{0, 2, 5, 10, 20, 30\}$. Note that the method with ``0" shared samples equals the baseline method \zxh{``DCIL w/ FedAvg"}. 
	
	We can observe that the larger shared dataset can perform better, while it also brings more communication costs and data privacy issues. Moreover, we can see that the test accuracy is prone to be saturated when the number of shared samples per class is larger than 20 in a local site, which is also the reason why we choose this parameter to carry out our experiments.
	
	\begin{table}[h!]
		\centering  
		\caption{Evaluating our method with the different numbers of shared samples per class on a local site with 5 session setting on subImageNet.}
		\label{ss}
		\renewcommand\arraystretch{1.2}
		\renewcommand\tabcolsep{2.5pt}
		\footnotesize
		\begin{tabular}{cccccccc}
			\hline\hline
			Number of & \multicolumn{6}{c}{Encountered Classes} & \multicolumn{1}{c}{Average} \\ \cline{2-7}
			Shared Samples & 50 & 60 & 70 & 80 & 90 & 100 & \multicolumn{1}{c}{Acc.} \\ \hline
			0  & 83.52 &70.83 &63.34 &58.25 &53.16 &50.14 &63.21  \\ 
			2  & 83.52 & 73.06&	67.14&	62.45&	58.13&	55.22&	66.70   \\
			5 & 83.52&	73.26&	68.35&	63.72&	59.04&	57.04&	67.48 \\
			10 & 83.52&	73.57&	68.28&	64.06&	60.23&	57.14& 67.80  \\
			\textbf{20} &83.52&	\textbf{73.43}& 	\textbf{68.91} &	\textbf{63.98}&\textbf{60.31} &	\textbf{57.92}&	\textbf{68.01}  \\
			30 & 83.52 &73.45&	69.13&	64.34&	60.62&	58.10&	68.19 \\ \hline\hline
		\end{tabular}
	\end{table}
	
	\vspace{2mm}
	\noindent \textbf{The effect of the number of local epochs}
	\vspace{2mm}
	
\begin{figure}[ht!]
		\includegraphics[width=0.495\textwidth]{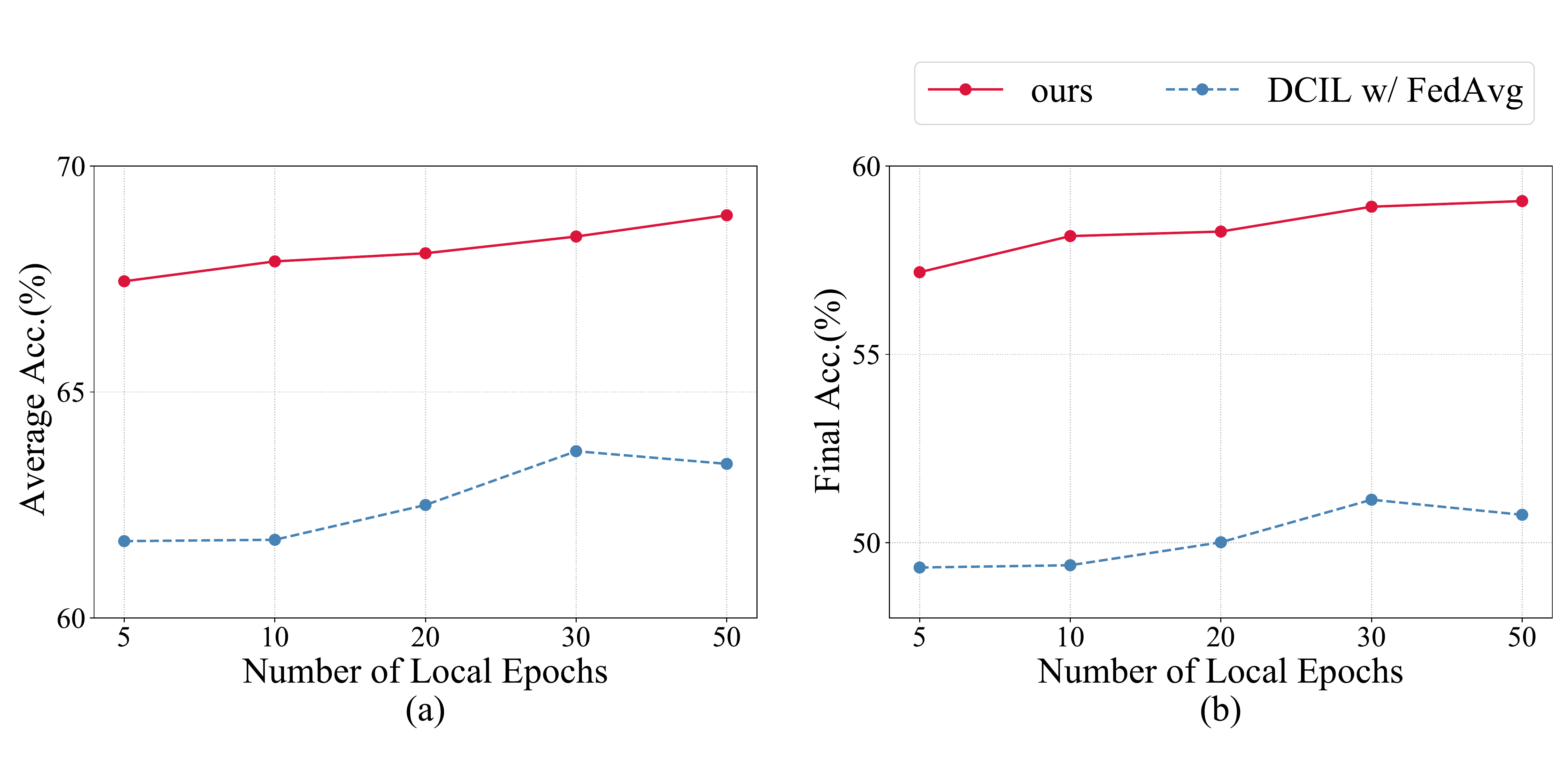}
		\caption{Comparison of average accuracy (a) and final accuracy (b) between our method DCID and baseline method ``DCID w/ FedAvg" with different numbers of local epochs.}
		\label{le}
	\end{figure}

	The number of the local epochs $E$ is a worth-exploring hyperparameter in distributed learning and fedrated learning, which affects the computation on local sites and the trade-off between communication cost and performance~\cite{lin2020ensemble}. Figure~\ref{le} compares the test accuracy of $E$ in the range of {\{5, 10, 20, 30, 50\}} with 300 total training epochs. The experiments use the LUCIR method with the 5-session and 5 local site settings.
	
	We can observe from Figure~\ref{le} that our DCID method consistently outperforms \zxh{the baseline method (``DCIL w/ FedAvg")} on both average accuracy and final accuracy for all numbers of local epochs.
	
	Specifically, we found that the performance of our method improves the average accuracy and final accuracy of the baseline method by $5.49\%$ and $8.17\%$ on average, respectively. Moreover, we observe that the longer the local training period takes, the better our method works. 
	The reason can be summarized that longer local training leads to higher quality of the ensemble and hence a better distillation result for the models~\cite{kuncheva2003measures}.
	In contrast, the performance of baseline method saturates and even degrades after $E$=30, which is consistent to the observations in previous literature~\cite{Mcmahan2016Communication,2020FederatedMA}. Fortunately, this phenomenon is alleviated in the proposed DCID.
	
	\vspace{2mm}
	\noindent \textbf{\zxhzxh{The effect of the number of anchors}}
	\vspace{2mm}
	
	\zxhzxh{
		The class incremental learning methods in our experiments all store several the old class anchors to represent the old knowledge. Though storing more anchors may helpful for the performance, it also brings more memory overhead and higher computation cost. Table~\ref{tab:anchor} reports the average accuracy achieved by using different numbers of anchors per class in a local site. It is observed that the test accuracy is prone to be saturated when the number of anchors per class is larger than 20. As the anchor number used in the original LUCIR~\cite{Hou_2019_CVPR} and TPCIL~\cite{tao_eccv20} algorithm is 20, which reaches a good balance of the accuracy and the memory costs, we set the number of anchors per class as 20 in the experiments.
	}
	
	\begin{table}[h!]
		\centering  
		\caption{Comparison of average accuracy with different numbers of anchors per class. }
		\label{tab:anchor}
		\renewcommand\arraystretch{1.2}
		\renewcommand\tabcolsep{5pt}
		\footnotesize
		\begin{tabular}{ccccccc}
			\hline\hline
			\multirow{2}{*}{Methods} & \multicolumn{4}{c}{Number of Anchors} \\ \cline{2-6}&   5      & 10        & 20       & 30   & 40 \\\hline
			Baseline   &  57.83   &   60.46  &   \textbf{63.21} & 64.18  & 64.67 \\
			DCID &62.79 & 65.52&\textbf{68.01} & 69.48& 69.71
			\\\hline\hline
		\end{tabular}
	\end{table}
	
	\vspace{2mm}
	\noindent \textbf{The effect of the number of local sites}
	\vspace{2mm}
	
	We further analyze the influences when our model communicates with the different numbers of local sites. In the experiments, we use LUCIR as the CIL method on subImageNet dataset with the 5 session setting. The number of local sites $M$ is in the range of $\{ 2, 5, 10, 20\}$. The anchor number is fixed to 100 per class. Table~\ref{tab:ls} shows the average accuracy and the final accuracy on the testing set achieved by different settings of local site number $M$.
	
	We can observe that with the increase of local site number~$M$, both the accuracies of the baseline and our method decrease, which is consistent with observations in~\cite{Mcmahan2016Communication,2020FederatedMA}. However, we can still see that our DCID achieves significantly better performance than the baseline in all settings consistently, which demonstrates the efficiency of our method. Moreover, with the increase of the number of local sites, our method has a greater improvement than the baseline method. This improvement is because that a larger number of  local sites enrich the diversity of models better and obtain higher-quality ensemble \zxhzxh{\zxhzxh{outputs}}~\cite{kuncheva2003measures}.

	\begin{table}[h!]
		\centering  
		\caption{Comparison of average accuracy and final accuracy using our method and baseline with different numbers of local sites. }
		\label{tab:ls}
		\renewcommand\arraystretch{1.2}
		\renewcommand\tabcolsep{5pt}
		\footnotesize
		\begin{tabular}{cccccc}
			\hline\hline
			& \multirow{2}{*}{Methods} & \multicolumn{4}{c}{Number of Local Sites} \\ \cline{3-6}&     & 2      & 5        & 10       & 20  \\\hline
			Average  & Baseline   & 67.13    & 63.21    & 62.86    & 61.71   \\
			Acc.   & DCID (Ours)                     & \textbf{70.07}    & \textbf{68.01}  & \textbf{67.61}    & \textbf{66.27}   \\\hline
			Final      & Baseline                 & 56.32    & 51.14    & 48.92    & 45.46   \\
			Acc.                    & DCID (Ours)                     & \textbf{61.66}    & \textbf{58.92}    & \textbf{56.54}    & \textbf{53.82}  \\\hline\hline
		\end{tabular}
	\end{table}
	
	\vspace{2mm}
	\noindent \textbf{\zxhzxh{The training time cost}}
	\vspace{2mm}
	
	\zxhzxh{We further evaluate the computational costs of typical class incremental learning (CIL) approaches and our decentralized class incremental learning (DCIL) approach, DCID. As both typical CIL and our proposed DCIL approaches generate a unified model, when using the same backbone, the inference time will be the same. As a result, we only have to compare the total training time.  }
	
	\zxhzxh{
		We have conducted experiments of total training time of four CIL algorithms and our DCID on CIFAR100 dataset with 5 local sites in one incremental learning session in Table~\ref{tab:time}. All training time cost is calculated from the start of training to convergence. The hyperparameters of four typical CIL methods are consistent with original paper. All the experiments are conducted on TITAN XP GPUs.}
	
	% 要不要强调？Note that in our experiments, the DID training processes of each local models are usually carried out synchronously. 
	\zxhzxh{
		Based on these results, we discuss the influences of decentralizing incremental learning in the computational costs. Typical incremental learning approaches train on a centralized dataset and generate one model. Instead, DCID consists of three processes: Decentralized Incremental Knowledge Distillation (DID), Decentralized Collaborative Knowledge Distillation (DCD) and Decentralized Knowledge Aggregated Distillation (DAD). Compared to typical class incremental learning approaches, the training data of DCID are distributed in different local sites. There are extra costs in model distribution and aggregation. Nevertheless, in each local site, we only have to train local models on a part of the dataset in the training process synchronously. As a result, the total training costs are reduced.}
	\begin{table}[h!]
		\centering  
		\caption{ Comparision of training time of typical CIL method and DICD of one session on CIFAR100.}
		\label{tab:time}
		\renewcommand\arraystretch{1.2}
		\renewcommand\tabcolsep{5pt}
		\footnotesize
		\begin{tabular}{ccccccc}
			\hline\hline
			{Methods} & iCARL&  LUCIR & ERDIL & TPCIL\\\hline
			Typical CIL  & 767s&   1041s  &   1094s &   1283s     \\
			DCID & \textbf{558s} & \textbf{662s} & \textbf{745s} & \textbf{871s}
			\\\hline\hline
		\end{tabular}
	\end{table}
	
	\zxhzxh{
		We can see that the training time of DCID is shorter than typical CIL methods, which mainly thanks to the fact that the local models can be trained simultaneously on the local dataset. Therefore, the proposed DCID framework is efficient.}

	\vspace{2mm}
	\noindent \textbf{\zxhzxh{The robustness to data variability}}
	\vspace{2mm}

	\zxhzxh{It is of great importance that the algorithms for machine learning are robust to potential data perturbations. 
		%as machine learning is applied on noisier and noisier data.
		To quantitatively validate the robustness of our method, inspired by the experiments in~\cite{8528557}, we perturb test images by jittering the hue of images and evaluate the performance against chromatic changes. We denote by $H$
		the hue of the original image, and by $\alpha_{h}$ the parameter of the magnitude of hue shift. The hue of image after processing is randomly sampled within the interval of $[H-\alpha_{h}, H+\alpha_{h}]$. We conducted experients on subImageNet with LUCIR method to investigate the performances of the baseline and our DCID with different $\alpha_{h}$.
		The results are shown in Table~\ref{tab:robust}.}
	
	\begin{table}[h!]
		\centering  
		\caption{Results with LUCIR by jittering the hue of test images on subImageNet. }
		\label{tab:robust}
		\renewcommand\arraystretch{1.2}
		\renewcommand\tabcolsep{5pt}
		\footnotesize
		\begin{tabular}{cccccc}
			\hline\hline
			& \multirow{2}{*}{Methods} & \multicolumn{4}{c}{$\alpha_h$} \\ \cline{3-6}&     & 0    & 0.1     &0.3    & 0.5\\\hline
			Average  & Baseline   & 63.21     & 61.67   & 58.11    & 56.23   \\
			Acc.   & DCID (Ours)                     & \textbf{68.01}    & \textbf{65.25}  & \textbf{60.40}    & \textbf{58.45}   \\\hline
			Final      & Baseline                 & 50.14   & 47.98  & 44.40   & 43.38  \\
			Acc.                    & DCID (Ours)                     & \textbf{57.92}    & \textbf{53.90}    & \textbf{50.06}    & \textbf{47.70}  \\\hline\hline
		\end{tabular}
	\end{table}

	\zxhzxh{Though the test accuracy is declined by the jittering of image hue, our DCID still outperforms the baseline method, proving the robustness of our method against chromatic perturbations.}
	
	%\zxh{Heterogeneous Distribution of Local Data}
	\subsection{\XP{Evaluation under the non-IID setting}}
	\label{subses:Heterogeneous}
	\vspace{2mm}
	
	\zxh{In data heterogeneously distributed settings, training data of new classes in session $t$ are distributed \XP{independently of class labels.} Traning data usually follows Dirichlet distribution over $L^{(t)}$ classes for non-IID setting in distributed learning and federated learning, as used in \cite{yurochkin2019bayesian,hsu2019measuring, lin2020ensemble}.} 
	
\begin{figure}[ht!]
		\includegraphics[width=0.495\textwidth]{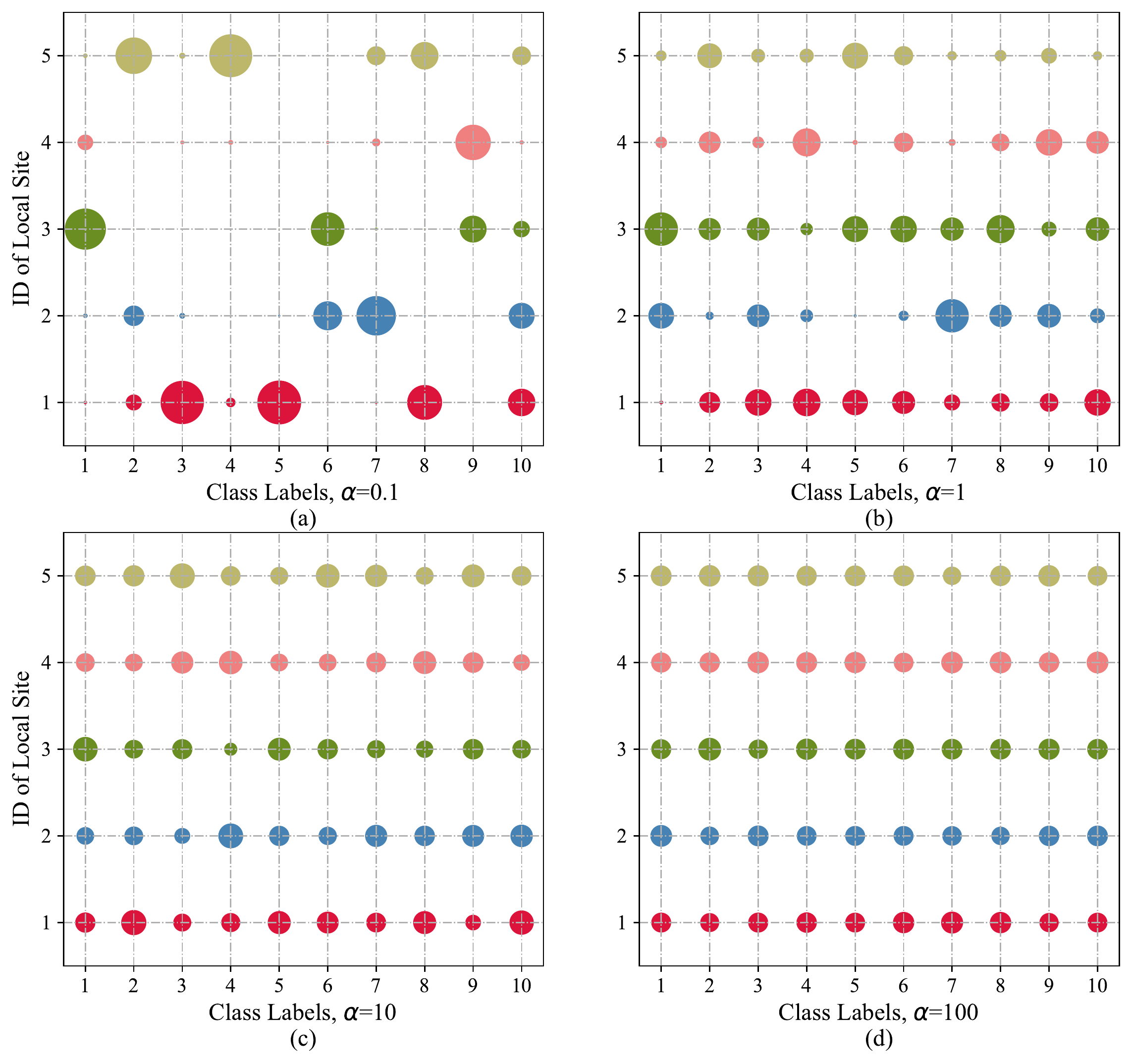}
		\caption{\zxh{Visualization of training samplers per class allocated to each local site in one session (5-session setting), for different $\alpha$ values of the Dirichlet distribution. $x$-axis indicates class labels and $y$-axis indicates IDs of local sites. The size of each dot reflects the magnitude of the samples number.}}
		\label{dirichlet}
	\end{figure}

	\zxh{Therefore, we follows the Dirichlet distribution $\alpha\sim\rm Dir(\alpha \bf p)$ to synthesize non-IID training data distributions in our experiments. The value of $\alpha$ is a concentration parameter controlling the degree of non-IID-ness among multiple local sites, while $\bf p$ characterizes a prior class distribution over $L^{(t)}$ classes in incremental session $t$. With $\alpha \rightarrow 0$, each local site holds training samples from only one random class; with $\alpha \rightarrow \infty$, all local sites have identical distributions to the prior class distribution. The prior class distribution is set to uniform distribution in our experiments. Therefore, a smaller $\alpha$ indicates higher data heterogeneity. In this work, we use the 5-session setting as an example, so each session contains 10 classes in both CIFAR100 and subImageNet. To better understand the local data distribution we considered in the experiments, we visualize the effects of adopting different $\alpha$ for 5-session setting of CIFAR100 and subImageNet in Figure~\ref{dirichlet}.}
	
\begin{figure}[ht!]
		\includegraphics[width=0.49\textwidth]{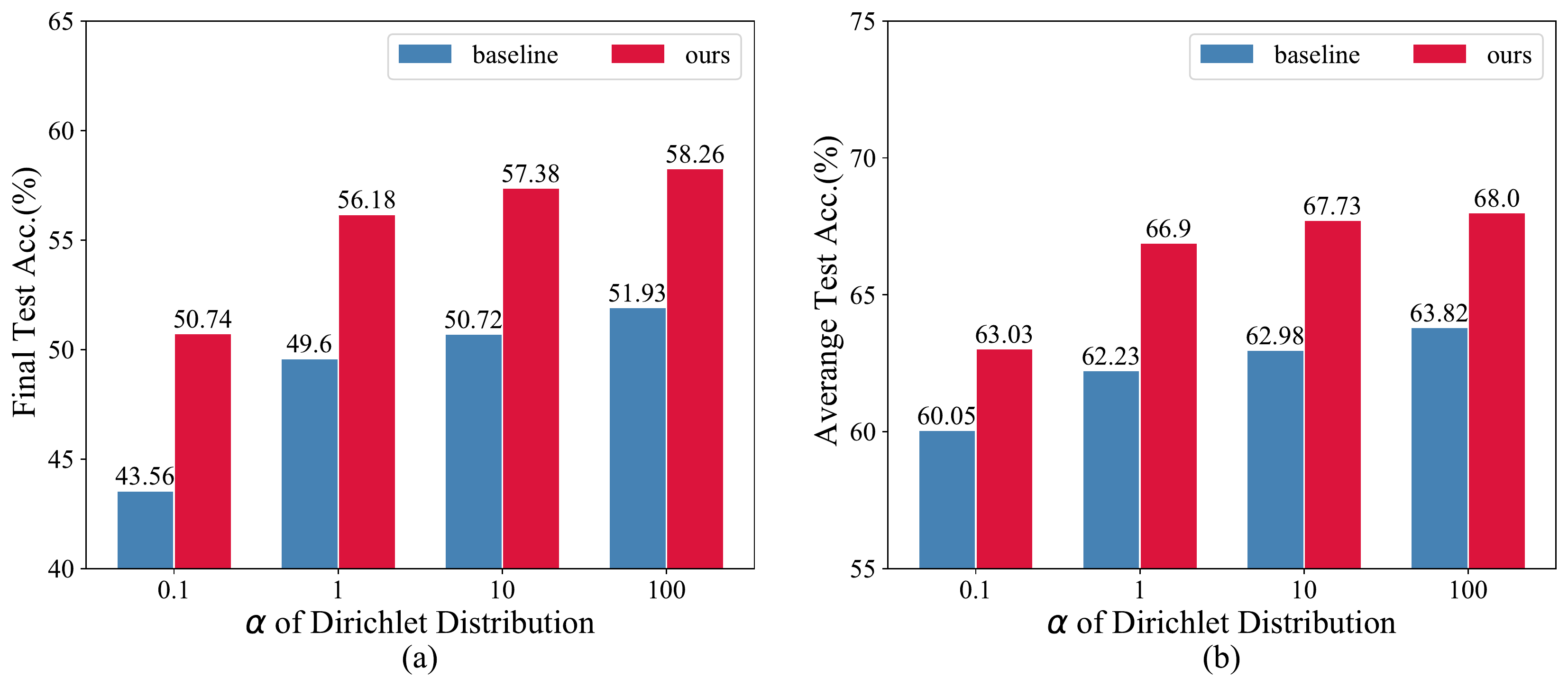}
		\caption{{Comparison of average accuracy (a) and final accuracy (b) between our method DCID and baseline method ``DCIL w/ FedAvg" using LUCIR w.r.t different $\alpha$ of Dirichlet distribution on subImageNet.}}
		\label{alpha}
	\end{figure}

\begin{figure}[ht!]
		\includegraphics[width=0.49\textwidth]{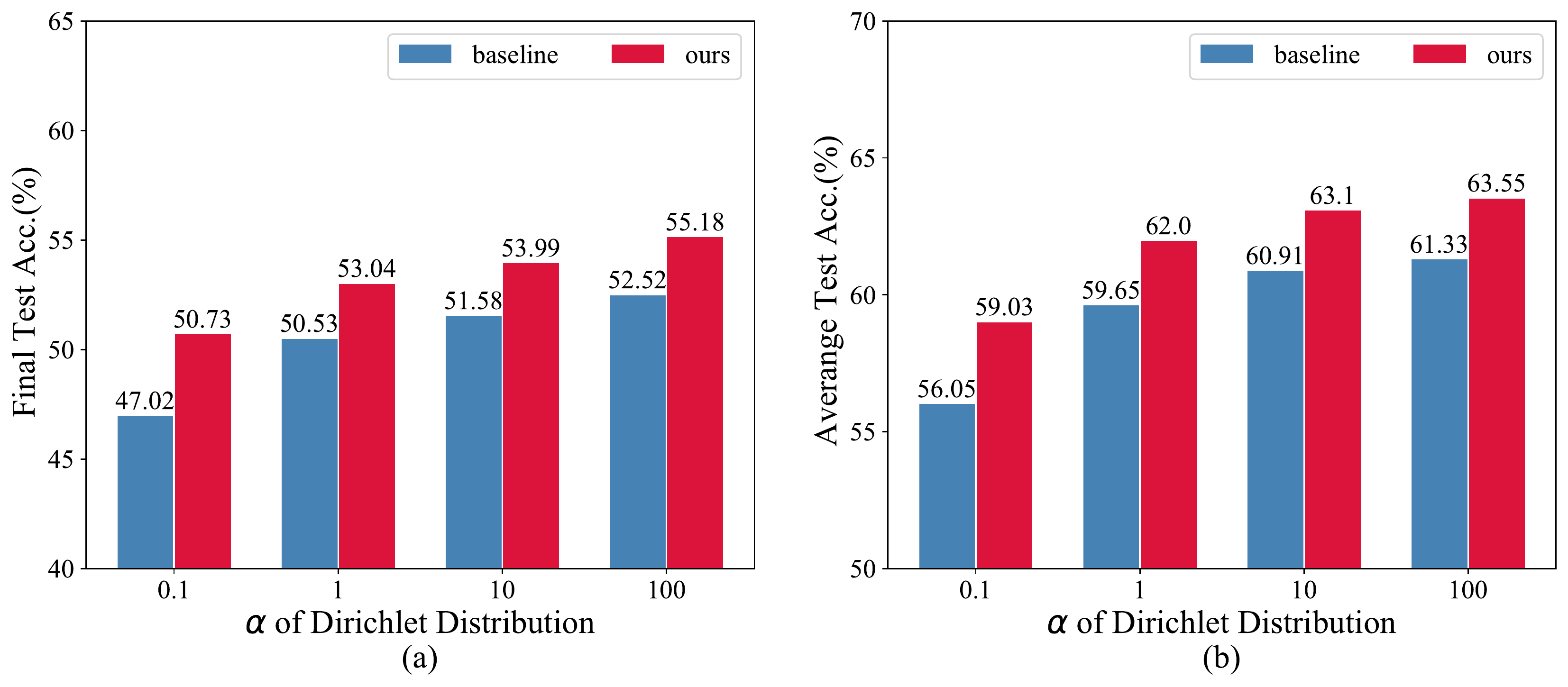}
		\caption{\zxh{Comparison of average accuracy (a) and final accuracy (b) between our method DCID and baseline method ``DCIL w/ FedAvg" using iCARL w.r.t different $\alpha$ of Dirichlet distribution on CIFAR100.}}
		\label{alpha_cifar}
	\end{figure}

	\zxh{
		We conduct experiments on $\alpha=0.1,1,10$ and $100$ on subImageNet with LUCIR and CIFAR100 with iCARL. We can observe that our DCID consistently outperforms baseline method ``DCIL w/ FedAvg", both in final accuracies and average accuracies, as shown in Figure~\ref{alpha} and~\ref{alpha_cifar}. %Table~\ref{tab:niid_i},\ref{tab:niid},\ref{tab:niid_t}. 
		It demonstrates the effectiveness of our proposed framework for heterogeneous distribution of local data. It's worth noting that the gain of DCID is still notable when the data distributions are highly heterogeneous (with a small $\alpha$).}
	
	\section{Conclusion}
	We initiate the study of decentralized deep incremental learning, which handles continuous streams of data coming from different sources.
	\XP{It is distinct to existing studies on (deep) incremental learning and distributed learning. Incremental learning can only update a model given a data stream coming from a single repository, while neither distributed learning nor federated learning can handle continuous data streams. The study of decentralized deep incremental learning is thus significant and challenging.}
	%Incremental learning works for a situation where data keeps coming from the \emph{same} source, while distributed learning and federated learning are targeted at the scenarios where data come from different sources but usually for \emph{specific, closed} tasks. 
	To facilitate this study, we establish a benchmark. We then propose a decentralized composite knowledge incremental distillation method, 
	which further outperforms the baseline method{s} consistently by a large margin under different \XP{IID and non-IID} settings. 
	In future, the proposed method will be applied to multi-robot systems.
	
	\section*{Acknowledgment}
	This work is funded by National Key Research and Development Project of China under Grant 2019YFB1312000 and by National Natural Science Foundation of China under Grant No. 62076195. The authors are also grateful to  Ms. Liangfei ZHANG and Ms. Yu LIU for their comments and discussions.

	\ifCLASSOPTIONcaptionsoff
	\newpage
	\fi
	
	
	
	%\bibliographystyle{IEEEtran}
	
	% \bibliography{ref}
	% \end{thebibliography}
	%
	
	{\small
		\bibliographystyle{IEEEtran}
		\bibliography{ref}
	}
	
	
	% \end{IEEEbiography}
	
	
	% % insert where needed to balance the two columns on the last page with
	% %\newpage
	
	
	
	%\vfill
	

\end{document}